\documentclass[10pt,twocolumn,letterpaper]{article}

\usepackage{cvpr}              

\usepackage{graphicx}
\usepackage{amsmath}
\usepackage{amssymb}
\usepackage{booktabs}
\usepackage{algorithm}
\usepackage{algorithmic}
\usepackage{makecell}
\usepackage{multirow}
\usepackage[table]{xcolor}

\definecolor{cvprblue}{rgb}{0.21,0.49,0.74}
\usepackage[pagebackref,breaklinks,colorlinks,allcolors=cvprblue]{hyperref}

\def\paperID{6393} 
\def\confName{CVPR}
\def\confYear{2026}

\title{Authorize-on-Demand: Dynamic Authorization \\ with Legality-Aware Intellectual Property Protection for VLMs}

\author{Lianyu Wang$^1$\footnotemark[1], \; Meng Wang$^{2,3}$\footnotemark[1], \; Huazhu Fu$^4$\footnotemark[2], \; Daoqiang Zhang$^1$\footnotemark[2] \\
\small{$^1$The Key Laboratory of Brain-Machine Intelligence Technology, Ministry of Education, China}\\ 
\small{$^2$Centre for Innovation and Precision Eye Health, Yong Loo Lin School of Medicine, NUS, Singapore}\\ 
\small{$^3$Department of Ophthalmology, Yong Loo Lin School of Medicine, NUS, Singapore}\\
\small{$^4$Institute of High Performance Computing, Agency for Science, Technology and Research, Singapore}
}

\begin{document}
\maketitle
\begin{abstract}

The rapid adoption of vision-language models (VLMs) has heightened the demand for robust intellectual property (IP) protection of these high-value pretrained models. Effective IP protection should proactively confine model deployment within authorized domains and prevent unauthorized transfers. However, existing methods rely on static training-time definitions, limiting flexibility in dynamic environments and often producing opaque responses to unauthorized inputs.
To address these limitations, we propose a novel dynamic authorization with legality-aware intellectual property protection (\textbf{AoD-IP}) for VLMs, a framework that supports authorize-on-demand and legality-aware assessment. AoD-IP introduces a lightweight dynamic authorization module that enables \textbf{flexible, user-controlled authorization}, allowing users to actively specify or switch authorized domains on demand at deployment time. This enables the model to adapt seamlessly as application scenarios evolve and provides substantially greater extensibility than existing static-domain approaches. In addition, AoD-IP incorporates a dual-path inference mechanism that jointly predicts input legality-aware and task-specific outputs. Comprehensive experimental results on multiple cross-domain benchmarks demonstrate that AoD-IP maintains strong authorized-domain performance and reliable unauthorized detection, while supporting user-controlled authorization for adaptive deployment in dynamic environments.

\end{abstract}   
\renewcommand{\thefootnote}{\fnsymbol{footnote}} 
\footnotetext[1]{L.~Wang and M.~Wang contributed equally to this work.} 
\footnotetext[2]{Corresponding author: H.~Fu and D.~Zhang.} 
\section{Introduction}
\label{sec:intro}

\begin{figure}[!t]
  \centering
   \includegraphics[width=1\linewidth,trim=210 130 210 100,clip]{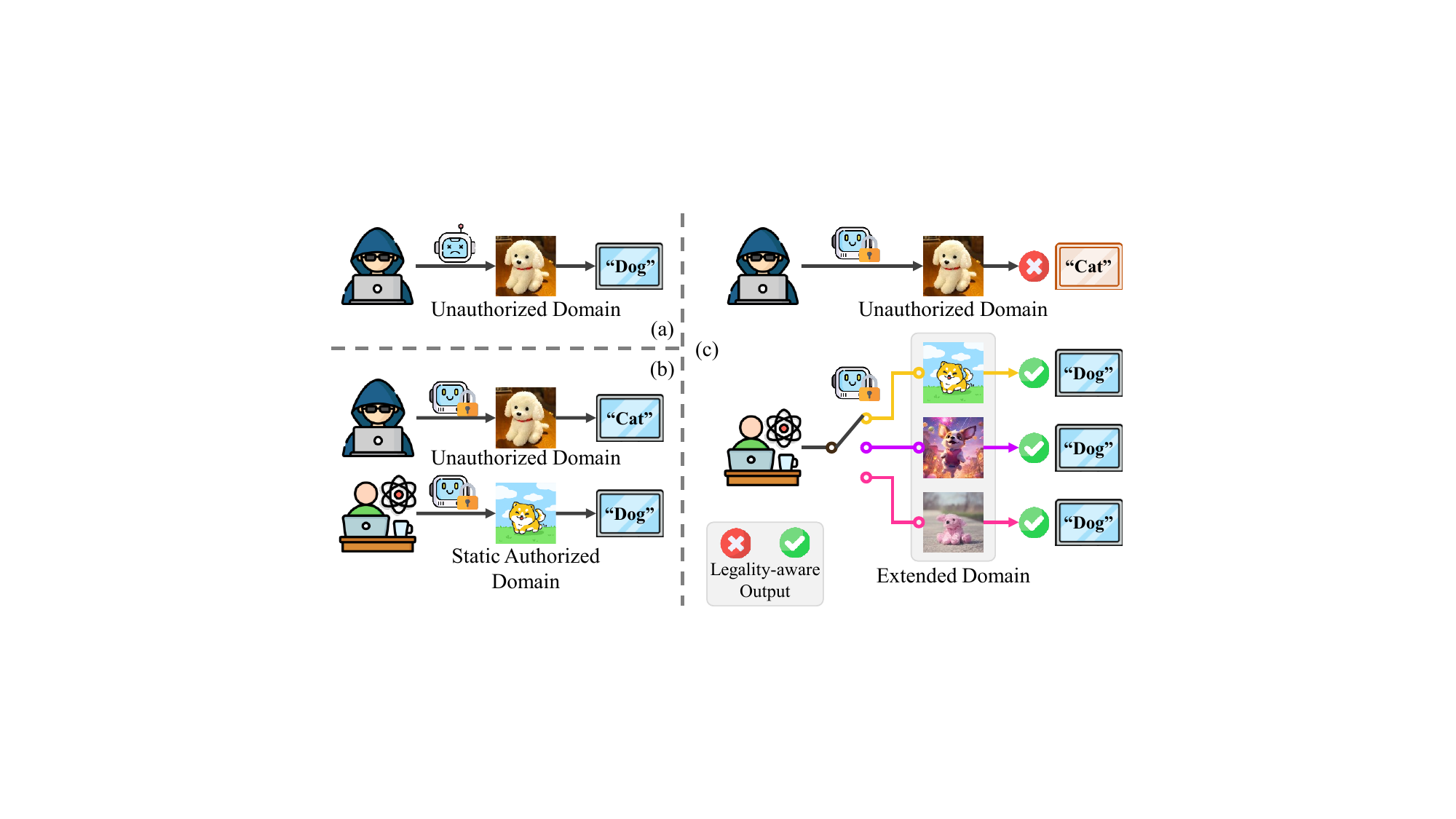}
   \caption{(a): Classical VLMs without IP protection; (b): Existing IP protection strategy (e.g., CUTI-Domain, CUPI-Domain) with static authorized domain; (c) The proposed AoD-IP allows users to actively specify or switch authorized domains on demand with legality-aware output.}
   \label{figure1}
\end{figure}

Deep learning models have become increasingly integral to industrial and commercial applications, making model intellectual property (IP) protection a critical concern. Vision-language models (VLMs), such as CLIP~\cite{CLIP}, which are trained on large-scale supervised datasets, encapsulate substantial computational resources~\cite{intro_device1,intro_device2,intro_device3}, extensive labeled or curated data~\cite{intro_data1,intro_data2,intro_data3}, and sophisticated architecture design~\cite{intro_arch1,intro_arch2}, representing significant investment by developers and organizations. These models are increasingly deployed in diverse real-world scenarios, such as autonomous driving~\cite{app_autodriving1}, medical image analysis~\cite{app_medical2}, industrial inspection~\cite{app_industrial1}, and financial forecasting~\cite{app_financial2}, where accurate and reliable predictions directly impact operational safety, business outcomes, and regulatory compliance.

Given their high economic and strategic value, proprietary models are particularly vulnerable to unauthorized use, domain transfer, or piracy. As illustrated in Fig.~\ref{figure1}(a), unauthorized parties may attempt to replicate, adapt, or fine-tune the models on new datasets without consent, potentially violating IP, undermining business competitiveness, or causing unsafe outcomes in critical applications. These risks are further amplified in cross-domain or multi-client deployments, where models may encounter unseen data distributions that can be exploited to extract features or knowledge. Consequently, robust and practical mechanisms for IP protection are essential, not only to safeguard the model developers’ investment but also to ensure secure deployment in dynamic real-world environments~\cite{IP1,IP2}.

To mitigate these risks, various IP protection strategies have been proposed, which can be categorized into ownership verification and applicability authorization methods.

Ownership verification aims to establish and prove model authorship~\cite{reflv,refbai,refpeng}. Typical approaches include model watermarking and fingerprinting. Model watermarking embeds a hidden signature into the model parameters or output behavior, allowing owners to verify their IP. Fingerprinting, on the other hand, identifies models based on unique behavioral patterns under specific inputs, enabling tampering detection and ownership verification. Although effective for authorship proof, watermarking and fingerprinting do not prevent unauthorized use or leakage of model performance in unauthorized domains. Both approaches primarily focus on post-hoc verification rather than active prevention of unauthorized usage.

Applicability authorization methods, in contrast, aim to control and restrict how models are used in practice~\cite{NTL,CUPI}. Representative techniques include non-transferable feature learning and isolation-domain frameworks such as CUTI-Domain, CUPI-Domain and NTL. Non-transferable learning encodes domain-specific features that are difficult to generalize to unauthorized domains, reducing the risk of feature piracy. Isolation-domain approaches create dedicated feature spaces to segregate authorized and unauthorized domains, limiting feature leakage. While these methods provide more active protection than ownership verification alone, existing frameworks face practical limitations. The static nature of authorized domain (as in Fig.~\ref{figure1}(b)) restricts model flexibility: new clients, data sources, or deployment conditions cannot be incorporated without retraining, which is computationally expensive. Meanwhile, uncontrolled outputs on unauthorized inputs raise safety and interpretability concerns, as models may output high-confidence but incorrect predictions, reducing user trust in real-world applications.

To overcome these limitations, as illustrated in Fig.~\ref{figure1}(c), we propose a novel Dynamic Authorization with Legality-Aware for VLMs Intellectual Property Protection (AoD-IP), which simultaneously enables \textbf{flexible, user-controlled authorization} and legality-aware verification. AoD-IP introduces a lightweight dynamic authorization module that allows the model to flexibly integrate new authorized domains on demand after initial training, eliminating the need for retraining and greatly enhancing flexibility in evolving deployment scenarios. In addition, AoD-IP incorporates a dual-path inference mechanism. Unlike conventional frameworks that output only task predictions, our approach jointly produces a legality-aware signal to indicate whether the input is authorized, together with the standard task-specific prediction. This dual-path formulation facilitates active monitoring, allowing users to distinguish between legitimate predictions and potential unauthorized usage. Furthermore, we design dedicated evaluation metrics to systematically assess the framework’s effectiveness in IP protection, ensuring comprehensive and quantifiable analysis of model performance in both authorized and unauthorized scenarios.
The main contributions of this work can be summarized as follows:

\begin{itemize}
\item{We propose \textbf{AoD-IP}, a novel framework for IP protection that integrates authorize-on-demand, effectively addressing both flexibility and security in real-world deployment.}
\item{We introduce a lightweight \textbf{dynamic authorization module} for post-training, user-controlled authorization, together with a \textbf{dual-path inference mechanism} that provides legality-aware verification alongside task prediction, improving flexibility under dynamic deployment conditions.}
\item{\textbf{Novel evaluation metrics} are designed to systematically assess the effectiveness and flexibility of IP protection frameworks.}
\item{Comprehensive experimental results on cross-domain benchmarks demonstrate that the proposed AoD-IP achieves competitive performance on IP protection tasks, highlighting its practical potential for secure and flexible deployment.\footnote[1]{https://github.com/LyWang12/AoD-IP}}
\end{itemize}
\section{Related Work}
\label{sec:related}

\subsection{Vision-Language Models and Parameter-Efficient Tuning}

Vision-Language Models (VLMs), such as CLIP~\cite{CLIP}, BLIP~\cite{BLIP}, FILIP~\cite{FILIP}, and GPT-4o~\cite{GPT-4o}, have demonstrated remarkable capabilities in learning joint visual–textual representations for multimodal understanding. By mapping visual and linguistic inputs into a shared semantic space, these models achieve strong zero-shot transfer and cross-domain generalization~\cite{DA,DG,pr}, driving their adoption in a wide range of real-world applications, including autonomous driving~\cite{app_autodriving1,app_autodriving2}, medical diagnosis~\cite{app_medical1,app_medical2}, industrial inspection~\cite{app_industrial1,app_industrial2}, and financial forecasting~\cite{app_financial1,app_financial2}.

To enhance the efficiency and adaptability of VLMs, parameter-efficient tuning techniques such as prompt tuning and adapter-based learning have been widely adopted. CoOp~\cite{CoOp} replaces handcrafted text prompts with learnable vectors while keeping the entire model fixed during training. CoCoOp~\cite{CoCoOp} further mitigates CoOp's overfitting issues by introducing a lightweight neural network that generates input-specific context tokens, thereby enhancing generalization. These approaches allow efficient adaptation without retraining the backbone, yet they also introduce security risks since models fine-tuned for particular domains may be transferred or misused, leading to potential IP leakage and unsafe deployment. As VLMs are increasingly deployed across diverse and sensitive domains, ensuring their operation within licensed or trusted boundaries has become a central challenge for secure and responsible AI deployment.

\subsection{Model Intellectual Property (IP) Protection}
The protection of deep learning model intellectual property (IP) has become increasingly important as models are deployed in sensitive or high-value industrial applications. Existing IP protection strategies can generally be categorized into two paradigms: \textit{ownership verification} and \textit{applicability authorization}.

Ownership verification aims to establish and prove model authorship, ensuring that a given model or its derivatives can be traced back to its legitimate owner. Typical approaches include model watermarking~\cite{watermarking1,watermarking2} and fingerprinting~\cite{fingerprint1,fingerprint2}.  

Model watermarking embeds hidden, verifiable signatures into model parameters, data, or outputs to enable ownership verification.
Lv~\textit{et al.}~\cite{reflv} proposed HufuNet, which embeds a pretrained encoder within the target network while privately retaining the decoder, allowing ownership to be verified through input–output consistency. Bai~\textit{et al.}~\cite{refbai} introduced BadCLIP, which employs trigger-aware prompts to jointly affect image and text encoders, showing that designed triggers can both verify ownership and detect backdoor behaviors.
Fingerprinting identifies models by their distinctive responses to specific inputs, enabling ownership verification and tampering detection without altering parameters.
Peng~\textit{et al.}~\cite{refpeng} proposed a contrastive fingerprinting scheme that perturbs the model and compares the resulting fingerprints with reference signatures for robust identification.
However, such methods mainly serve post-hoc verification, once a model has been leaked or fine-tuned, these traces become unreliable, limiting their effectiveness for active IP protection.

Applicability authorization methods focus on actively controlling and restricting how a model can be used in practice, particularly across different domains. Representative techniques include non-transferable feature learning and isolation-domain frameworks. Non-transferable learning constrains representations to domain-specific features that fail to generalize to unauthorized inputs, mitigating feature piracy. 
Wang~\textit{et al.}~\cite{NTL} employed kernel-based feature estimators to amplify domain distinctions and suppress transferable components, while Zeng~\textit{et al.}~\cite{refzeng} extended the paradigm to natural language processing using auxiliary domain classifiers.
Hong~\textit{et al.}~\cite{HNTL} introduced HNTL, leveraging causal disentanglement of content and style, and Deng~\textit{et al.}~\cite{SOPHON} introduced a meta-learning-based optimization strategy that confines the model within a local optimum of the authorized domain.
Isolation-domain frameworks explicitly separate the feature spaces of authorized and unauthorized domains to minimize cross-domain feature leakage and enforce domain isolation. Wang~\textit{et al.}~\cite{CUTI,CUPI} proposed a series of isolation-domain frameworks, including the CUTI-Domain~\cite{CUTI} and CUPI-Domain~\cite{CUPI}, which construct hierarchical feature spaces dedicated to authorized and unauthorized domains. By explicitly separating these domains within the representational space, such methods effectively suppress cross-domain feature leakage and performance transfer. However, when the authorized domain changes according to the practical requirements, these methods generally require retraining from scratch, leading to high computational and deployment costs.
In summary, while ownership verification ensures post-hoc authorship and applicability authorization restricts cross-domain misuse, both lack proactive flexibility. In contrast, AoD-IP introduces a dynamic authorization mechanism orthogonal to traditional security concerns (e.g., credential extraction, reverse engineering or replay attacks), thereby fortifying IP protection within evolving deployment environments.

\section{Method}

\begin{figure*}[!t]
\centering
\includegraphics[width=1\linewidth,trim=90 85 90 85,clip]{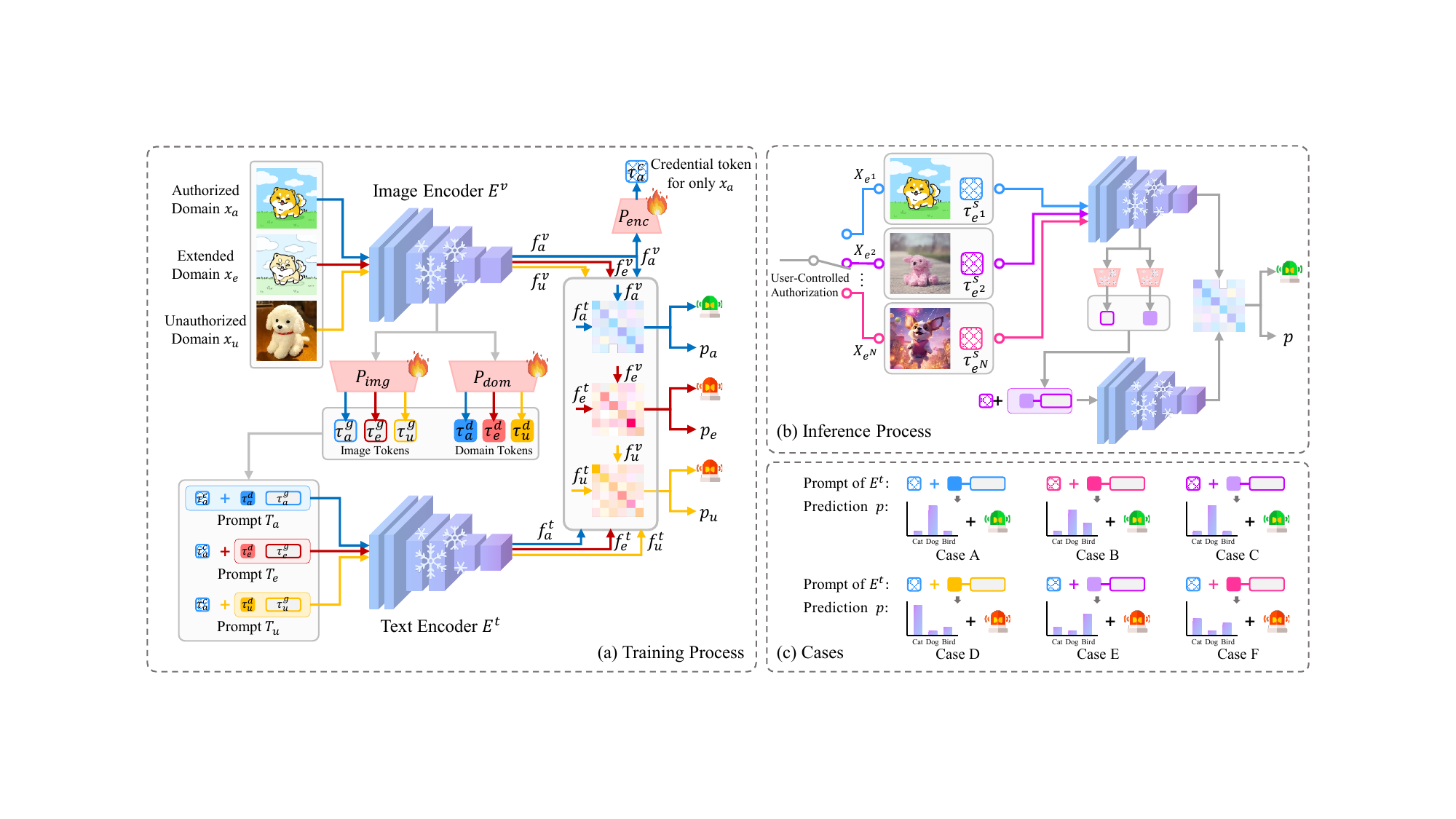}
\caption{(a) During training, authorized data ($x_a$), extended data ($x_e$), and unauthorized data ($x_u$) are simultaneously processed by the frozen CLIP visual encoder $E^v$ to extract visual features ($f_a^v$, $f_e^v$, $f_u^v$). The image projector $P_\mathrm{img}$ and domain projector $P_\mathrm{dom}$ generate image tokens ($\tau_a^g, \tau_e^g, \tau_u^g$) and domain-discriminative tokens ($\tau_a^d, \tau_e^d, \tau_u^d$) for the three domains, respectively.
In parallel, an encryption projector $P_\mathrm{enc}$ produces a credential token $\tau_a^c$ for authorized data. These tokens are concatenated and fed into the frozen text encoder $E^t$, producing the corresponding textual features ($f_a^t$, $f_e^t$, $f_u^t$). The final prediction $p$ is derived from the similarity between visual features ($f^v$) and textual features ($f^t$), while an auxiliary output path verifies the legitimacy of each prediction. Frozen modules are indicated by snowflakes, and trainable modules by spark markers. (b) During inference, users may request additional credential tokens from the model owner, which function as “domain keys.” By selecting or switching these keys, users can dynamically control which domain is activated, thereby obtaining valid predictions accordingly. (c) Several inference cases are shown: only matching credentials and inputs lead to valid outputs (e.g., Cases A-C), while mismatched inputs result in invalid predictions and trigger security alerts (e.g., Cases D-F).
}
\label{fig2}
\end{figure*}

\subsection{Problem Definition and Authorize-on-Demand Formulation for IP Protection}
\label{sec_definition}
The goal of IP protection is to confine a model’s recognition capability to its authorized domain while suppressing effectiveness on unauthorized data, thereby enforcing domain-specific recognition boundaries.

\textbf{Definition 1. Model IP Protection:}
\textit{Let $D_a={(x_{ai}, y_{ai})}_{i=1}^{N_a}$ denote the dataset of the authorized domain, and $D_u={(x_{ui}, y_{ui})}_{i=1}^{N_u}$ denote that of the unauthorized domain, where $N_a$ and $N_u$ represent the respective numbers of samples. Although $X_a$ and $X_u$ follow distinct data distributions, they share the same label space $Y$. The learning objective is to enable the model to correctly associate samples from the authorized domain with their labels while suppressing the transfer of this capability to unauthorized domains.}
\begin{equation}
\label{eq1}
D_a \perp D_u, \quad F(X_a) \rightarrow Y, \quad F(X_u) \perp Y,
\end{equation}
\textit{where $\perp$ indicates statistical independence.} 

However, existing IP protection methods are typically built upon a predefined and static authorized domain. When downstream user requirements change, these methods often require the model to be retrained from scratch, which is highly resource-intensive. To overcome this limitation, we propose an extensible task that enables flexible user-controlled authorization, formally defined as follows:

\textbf{Definition 2. Authorize-on-Demand Model IP Protection:}  
\textit{
Let the authorized / unauthorized domain be denoted as $D_a$ / $D_u$. The set of extended domains is defined as $D_e=\{D_{e^1}, D_{e^2}, \dots, D_{e^N}\}$, where $D_{e^n}={(x_{e^ni}, y_{e^ni})}_{i=1}^{N_{e^n}}$, and $N$ denotes the number of extended domains. All domains share the same label space $Y$ but differ in their data distributions. The goal of authorize-on-demand IP protection is to preserve the model’s performance on the authorized domain and to enable seamless replacement of $D_a$ by any $D_{e^n}$ without retraining, while maintaining isolation from the unauthorized domain. The task is formally constrained as follows:}
\begin{equation}
\label{eq:extensible}
\begin{split}
&D_a \perp D_e \perp D_u, \quad \mathcal{S}=\{a,e^1,e^2,\dots,e^N\}, \\
&F(X_k) \rightarrow Y, \text{where } k \text{is user-selected from } \mathcal{S} ,\\
&F(X_u) \perp Y.
\end{split}
\end{equation}

\subsection{Overview of the AoD-IP Architecture}
\label{sec_overview}
The overall architecture of AoD-IP is illustrated in Fig.~\ref{fig2}(a).
The input consists of three parallel components: authorized domain data $x_a$, extended domain data $x_e$ (detailed in Section~\ref{sec_extended}), and unauthorized domain data $x_u$.
These inputs are first processed by the frozen CLIP visual encoder $E^v$ to obtain deep visual representations, denoted as $f^v = [f^v_a, f^v_e, f^v_u]$.
The learnable image projector $P_\mathrm{img}$ and domain projector $P_\mathrm{dom}$ generate the image tokens $(\tau_a^g, \tau_e^g, \tau_u^g)$ and domain-discriminative tokens $(\tau_a^d, \tau_e^d, \tau_u^d)$ for the three domains, respectively.
Meanwhile, the encryption projector $P_\mathrm{enc}$ produces a unique credential token $\tau_a^c$ for the authorized domain (Section~\ref{sec_DAM}).
Subsequently, these three types of tokens are concatenated and passed through the frozen CLIP text encoder $E^t$ to obtain deep textual features $f^t = [f^t_a, f^t_e, f^t_u]$.
Finally, the similarity between $f^v$ and $f^t$ is computed to derive the final prediction $p$ and verify its legitimacy (Section~\ref{sec_output}).

\subsection{Design of Extended Domain}
\label{sec_extended}
In the AoD-IP framework, the model needs to establish a one-to-one correspondence between the authorized domain and its credential token $\tau_a^c$, ensuring that $\tau_a^c$ is valid only within the authorized domain (\textbf{Case A} of Fig.~\ref{fig2}(c)).
To prevent unauthorized activation, the $\tau_a^c$ becomes invalid for all other domains (\textbf{Cases D-F}), thereby safeguarding the exclusivity of authorized access.

To support this mechanism, an extended domain $x_e$ is introduced to simulate diverse and unknown domains that may emerge in real-world scenarios. It serves two complementary purposes:
(1) simulate diverse unknown domains with varying sources and styles from real-world scenarios; and
(2) proactively model potential future extended-authorized domains after training.

In practice, $x_e$ is generated by applying random style perturbations~\cite{randaugment} to the authorized domain, enriching domain diversity without introducing external data. We adopt these perturbations to deliberately simulate "hard-to-distinguish" shifts, as subtle domain discrepancies are inherently more challenging to isolate in the latent space. This approach enforces a robust authorization boundary using only existing data. When a new domain is officially authorized (\textbf{Cases B-C}), AoD-IP enables flexible, user-controlled domain switching through lightweight credential updates, eliminating the need for backbone retraining.

\subsection{Dynamic Authorization Module}
\label{sec_DAM}
The dynamic authorization module consists of three lightweight projectors: the image projector $P_\mathrm{img}$, the domain projector $P_\mathrm{dom}$, and the encryption projector $P_\mathrm{enc}$. First, the multi-scale features of each domain are extracted using the image encoder $E^v$ and concatenated before being fed into $P_\mathrm{img}$ and $P_\mathrm{dom}$. For the authorized, extended, and unauthorized domains, $P_\mathrm{img}$ generates the corresponding image tokens $(\tau_a^g, \tau_e^g, \tau_u^g)$, while $P_\mathrm{dom}$ produces the domain tokens $(\tau_a^d, \tau_e^d, \tau_u^d)$. The encryption projector $P_\mathrm{enc}$ receives the deep feature representation $f_a^v$ from the authorized domain and outputs its unique credential token $\tau_a^c$, which serves as a domain key to activate the corresponding authorized domain.

Subsequently, the authorized domain prompt $T_a$ is constructed by concatenating the three authorized tokens, defined as:
\begin{equation}
T_a = [\tau_a^c, \tau_a^g, \tau_a^d].
\end{equation}
This prompt is then fed into the frozen text encoder $E^t$ to generate textual features $f_a^t$. By comparing $f_a^v$ and $f_a^t$, the model produces the task prediction $p_a$ together with a legality-aware output reflecting whether the input is authorized.

For the extended and unauthorized domains, text prompts are constructed to simulate real-world unauthorized or mismatched domain conditions. Since adversaries do not possess a valid credential token corresponding to their input domain data, two possible situations occur:
(1) \textbf{Token missing}: the credential token is absent, yielding an incomplete token set and halting inference; or
(2) \textbf{Token misuse}: an existing credential token is improperly paired with non-authorized domain tokens in an attempt to illegally activate the model.
We instantiate case (2) as:
\begin{equation}
T_e = [\tau_a^c, \tau_e^g, \tau_e^d], \quad
T_u = [\tau_a^c, \tau_u^g, \tau_u^d].
\end{equation}
$T_e$ and $T_u$ are fed into the text encoder $E_t$ to generate textual representations $f_e^t$ and $f_u^t$, which are then compared with their corresponding visual features $f_e^v$ and $f_u^v$ to compute similarity scores. This process produces a dual-path output comprising the task predictions ($p_e$, $p_u$) and a legality-aware signal that identifies unauthorized inputs.
Under this setting, the mismatch between the input domain and the credential token effectively simulates unauthorized access, allowing AoD-IP to distinguish legitimate from illegitimate predictions.

\subsection{Dual-path Output}
\label{sec_output}
AoD-IP adopts a dual-path output mechanism that jointly produces task predictions and legality-aware output.
For each domain $i \in \{a,e,u\}$, the model outputs a prediction vector $p_i \in \mathbb{R}^{N+1}$ by computing similarity between the visual feature $f^v_i$ and the textual feature $f^t_i$. The first $N$ entries correspond to the task classes, and the last entry represents the unauthorized class. The legality-aware output $r_i$ is then defined as
\begin{equation}
r_i =
\begin{cases}
1, & \text{if } \arg\max(p_i) \neq C_{\mathrm{unauth}}, \\
0, & \text{if } \arg\max(p_i) = C_{\mathrm{unauth}},
\end{cases}
\end{equation}
where $C_{\mathrm{unauth}}$ denotes the $N+1$-th class representing unauthorized inputs. Through this dual-path design, AoD-IP can simultaneously determine what the input represents (task prediction) and whether it is authorized (legality-aware verification), ensuring secure inference within each domain.

\subsection{Training and Inference}
\label{sec_legitimacy}
To achieve secure and flexible model IP protection, AoD-IP adopts a unified training strategy. The overall training objective is formulated as:
\begin{equation}
\mathcal{L} = \mathcal{L}_{\mathrm{ce}}^{a} - \lambda_1 \cdot \mathcal{L}_{\mathrm{ce}}^{a \rightarrow u} + \mathcal{L}_{\mathrm{ce}}^{u} + \mathcal{L}_{\mathrm{ce}}^{e} - \mathcal{L}_{\mathrm{kl}}.
\end{equation}

The classification loss $\mathcal{L}_{\mathrm{ce}}^{a}$ ensures accurate task prediction within the authorized domain:
\begin{equation}
\mathcal{L}_{\mathrm{ce}}^{a}
= \lambda_1 \cdot \mathcal{L}_{\mathrm{ce}}(p_a, y_a),
\end{equation}
where $y_a$ denotes the ground-truth label corresponding to the authorized domain. The second term penalizes the case where authorized samples are misclassified as unauthorized:
\begin{equation}
\mathcal{L}_{\mathrm{ce}}^{a \rightarrow u}
= \mathcal{L}_{\mathrm{ce}}(p_a, y_{N+1}),
\end{equation}
where $\lambda_1$ is empirically set to $0.1$ to balance discrimination and stability.

$\mathcal{L}_{\mathrm{ce}}^{u}$ and $\mathcal{L}_{\mathrm{ce}}^{e}$ guide the model to classify samples into the “unauthorized” category, thereby suppressing knowledge transfer to unauthorized domains:
\begin{equation}
\mathcal{L}_{\mathrm{ce}}^{u}
= \mathcal{L}_{\mathrm{ce}}(p_u, y_u),\quad \mathcal{L}_{\mathrm{ce}}^{e}
= \mathcal{L}_{\mathrm{ce}}(p_e, y_e).
\end{equation}

Finally, the Kullback–Leibler divergence~\cite{van2014renyi} is introduced to enhance inter-domain feature separability and prevent feature overlap between the authorized and extended domains:
\begin{equation}
\mathcal{L}_{\mathrm{kl}} = \mathrm{KL}(f_a^t \parallel f_e^t).
\end{equation}

During inference, AoD-IP retains only the frozen backbone and shared modules, including the visual encoder $E^v$, text encoder $E^t$, projector $P_{\mathrm{img}}$, and $P_{\mathrm{dom}}$, as in Fig.~\ref{fig2}(b).
The encryption projector $P_{\mathrm{enc}}$ is securely maintained by the model owner and is not publicly released. 

Each inference input consists of a data sample and its credential token.
If the credential matches the data, the model outputs both the task prediction and the legality-aware output (Fig.~\ref{fig2}(c), \textbf{Case A}); otherwise, the result is flagged as “unauthorized,” as illustrated in \textbf{Cases D–F}.
When a new domain is introduced on demand after training, users can obtain additional credential tokens generated by $P_{\mathrm{enc}}$ from the model provider.
These credentials enable seamless, user-controlled domain switching without retraining the backbone, acting like keys that unlock the corresponding domains (\textbf{Cases B–C}), they support flexible deployment under dynamic authorization settings.
\section{Experiment}
\subsection{Implementation Details}

We comprehensively evaluate the effectiveness of the proposed AoD-IP framework against SOTA methods~\cite{NTL,CUTI,CUPI,HNTL,SOPHON,IPCLIP} on multiple public benchmarks:
\begin{itemize}
\item \textbf{Office-31}~\cite{office31} contains images from three domains, namely Amazon (Am), Dslr (Ds), and Webcam (We), covering 31 object categories with more than 4,000 images.
\item \textbf{Office-Home-65}~\cite{home} consists of four visually distinct domains, including Art (Ar), Clipart (Cl), Product (Pr), and Real-World (Re), spanning 65 categories and over 15,000 images.
\item \textbf{Mini-DomainNet}~\cite{mini} includes four diverse domains, namely Clipart (Cl), Painting (Pa), Real (Re), and Sketch (Sk), with 126 categories and more than 140,000 images.
\end{itemize}
All experiments are implemented using the PyTorch framework on an NVIDIA GeForce RTX 3090 GPU (24 GB memory). The proposed AoD-IP adopts the pretrained CLIP backbone as its visual-language encoder. For a fair comparison, all comparison methods are re-implemented as VLM-based variants according to their original source codes. Following standard evaluation protocols, classification accuracy (\%) is used as the primary performance metric. To provide a comprehensive assessment of IP protection effectiveness, the following additional metrics are also employed:
\begin{itemize}
    \item Authorized domain accuracy Drop: $Drop_a = A_a^{sl} - A_a^{ip}$, where $A_a^{sl}$ / $A_a^{ip}$ denotes the accuracy of model without/with IP protection on the authorized domain. A smaller $Drop_a$ indicates less negative performance impact caused by IP protection on the authorized domain.
    \item Unauthorized domain accuracy drop: $Drop_u = A_u^{sl} - A_u^{ip}$, where a larger $Drop_u$ reflects stronger suppression of model performance on the unauthorized domain.
    \item Weighted drop difference: $W_{u-a} = A_a^{ip} \cdot (Drop_u - Drop_a)$, which jointly balances the trade-off between maintaining task performance on authorized domains and suppressing unauthorized generalization.
    \item Accuracy cross drop: $D_{u-a} = A_a^{ip} \cdot [A_a^{ip} - A_u^{ip}]$, which jointly balances the accuracy on authorized domain and the accuracy gap between domains.
    \item Legality discrimination accuracy: We introduce $R_a$, $R_e$, $R_u$ to quantify the accuracy of legality-aware output.
\end{itemize}

\begin{table}[!t]
\renewcommand\arraystretch{1}
\setlength{\tabcolsep}{6pt}
  \centering
      \caption{The results of target-specified AoD-IP on the Office-31~\cite{office31}.}
  \resizebox{0.48\textwidth}{!}{
\begin{tabular}{cc|cc|ccc}
    \toprule
    $X_a \rightarrow X_u$ & $A^{ip}_a \rightarrow A^{ip}_u$ & $X_{e^1}$ & $A^{ip}_{e^1} \rightarrow A^{ip}_u$ & $R_a$ & $R_e$ & $R_u$ \\
    \midrule
    Am $\rightarrow$ Ds & 78.8 $\rightarrow$ 0.0 & We & 97.5 $\rightarrow$ 0.0 &  100.0 & 98.5 &  100.0 \\
    Am $\rightarrow$ We & 79.5 $\rightarrow$ 0.0 & Ds & 95.1 $\rightarrow$ 2.5 &  100.0 & 96.3 & \ 98.8 \\
    Ds $\rightarrow$ Am & 95.7 $\rightarrow$ 0.0 & We & 94.3 $\rightarrow$ 0.0 & \ 97.5 & 97.5 & \ 99.4 \\
    Ds $\rightarrow$ We & 97.2 $\rightarrow$ 0.0 & Am & 79.0 $\rightarrow$ 0.0 & \ 98.8 & 97.5 &  100.0 \\
    We $\rightarrow$ Am & 93.1 $\rightarrow$ 0.0 & Ds & 94.6 $\rightarrow$ 1.3 &  100.0 & 98.8 & \ 98.8 \\
    We $\rightarrow$ Ds & 92.6 $\rightarrow$ 0.0 & Am & 80.3 $\rightarrow$ 0.0 &  100.0 & 96.3 &  100.0 \\
    \bottomrule
  \end{tabular}}
  \label{ts_office_detail}
\end{table}

\begin{table}[!t]
\renewcommand\arraystretch{1}
\setlength{\tabcolsep}{4pt}
  \centering
      \caption{The results of target-specified AoD-IP on Office-Home-65~\cite{home}.}
  \resizebox{0.48\textwidth}{!}{
  \begin{tabular}{cc|cccc|ccc}
    \toprule
    $X_a \rightarrow X_u$ & $A^{ip}_a \rightarrow A^{ip}_u$ & $X_{e^1}$ & $A^{ip}_{e^1} \rightarrow A^{ip}_u$ & $X_{e^2}$ & $A^{ip}_{e^2} \rightarrow A^{ip}_u$ & $R_a$ & $R_e$ & $R_u$ \\
    \midrule
    Ar $\rightarrow$ Cl & 84.8 $\rightarrow$ \ 2.8 & Pr & 92.3 $\rightarrow$ \ 5.0 & Re & 89.5 $\rightarrow$ \ 6.3 & 93.8 & 98.1 & 94.6 \\
    Ar $\rightarrow$ Pr & 86.3 $\rightarrow$ \ 4.3 & Cl & 75.3 $\rightarrow$ \ 2.0 & Re & 89.8 $\rightarrow$  18.0 & 99.0 & 95.9 & 91.1 \\
    Ar $\rightarrow$ Re & 86.3 $\rightarrow$  14.3 & Cl & 77.0 $\rightarrow$ \ 8.5 & Pr & 92.8 $\rightarrow$  23.8 & 91.5 & 98.9 & 83.9 \\
    Cl $\rightarrow$ Ar & 75.5 $\rightarrow$ \ 8.0 & Pr & 92.8 $\rightarrow$ \ 5.5 & Re & 89.3 $\rightarrow$  14.5 & 93.8 & 95.6 & 86.1 \\
    Cl $\rightarrow$ Pr & 73.5 $\rightarrow$ \ 1.8 & Ar & 85.0 $\rightarrow$ \ 4.8 & Re & 90.8 $\rightarrow$ \ 7.5 & 96.3 & 94.6 & 95.1 \\
    Cl $\rightarrow$ Re & 78.0 $\rightarrow$ \ 5.8 & Ar & 85.8 $\rightarrow$  10.8 & Pr & 92.3 $\rightarrow$  11.3 & 93.8 & 92.8 & 90.3 \\
    Pr $\rightarrow$ Ar & 92.3 $\rightarrow$ \ 4.8 & Cl & 75.3 $\rightarrow$ \ 4.0 & Re & 91.0 $\rightarrow$  11.5 & 97.5 & 96.0 & 90.3 \\
    Pr $\rightarrow$ Cl & 93.5 $\rightarrow$ \ 2.3 & Ar & 83.8 $\rightarrow$ \ 2.0 & Re & 89.5 $\rightarrow$ \ 1.5 & 99.3 & 95.8 & 97.7 \\
    Pr $\rightarrow$ Re & 92.0 $\rightarrow$  15.8 & Ar & 85.3 $\rightarrow$  19.5 & Cl & 72.5 $\rightarrow$ \ 4.3 & 99.0 & 92.5 & 85.4 \\
    Re $\rightarrow$ Ar & 89.8 $\rightarrow$  24.3 & Cl & 72.3 $\rightarrow$ \ 3.3 & Pr & 94.0 $\rightarrow$ \ 4.0 & 90.8 & 97.4 & 88.7 \\
    Re $\rightarrow$ Cl & 90.0 $\rightarrow$ \ 4.0 & Ar & 85.8 $\rightarrow$ \ 3.3 & Pr & 91.0 $\rightarrow$ \ 5.5 & 97.0 & 96.6 & 95.2 \\
    Re $\rightarrow$ Pr & 89.5 $\rightarrow$  18.5 & Ar & 85.5 $\rightarrow$ \ 3.5 & Cl & 75.0 $\rightarrow$ \ 6.3 & 92.3 & 94.8 & 89.2 \\
    \bottomrule
  \end{tabular}}
  \label{ts_home_detail}
\end{table}

\begin{table}[!t]
\renewcommand\arraystretch{1}
\setlength{\tabcolsep}{4pt}
  \centering
      \caption{The results of target-specified AoD-IP on the Mini-DomainNet~\cite{mini}.}
  \resizebox{0.48\textwidth}{!}{
  \begin{tabular}{cc|cccc|ccc}
    \toprule
    $X_a \rightarrow X_u$ & $A^{ip}_a \rightarrow A^{ip}_u$ & $X_{e^1}$ & $A^{ip}_{e^1} \rightarrow A^{ip}_u$ & $X_{e^2}$ & $A^{ip}_{e^2} \rightarrow A^{ip}_u$ & $R_a$ & $R_e$ & $R_u$ \\
    \midrule
    Cl $\rightarrow$ Pa & 84.0 $\rightarrow$  11.9 & Re & 90.9 $\rightarrow$  12.6 & Sk & 81.3 $\rightarrow$  12.1 & 94.1 & 95.7 & 86.4 \\
    Cl $\rightarrow$ Re & 84.4 $\rightarrow$  13.6 & Pa & 80.5 $\rightarrow$  12.2 & Sk & 81.6 $\rightarrow$ \ 8.6 & 91.2 & 91.0 & 88.1 \\
    Cl $\rightarrow$ Sk & 84.4 $\rightarrow$  17.7 & Pa & 81.9 $\rightarrow$ \ 8.2 & Re & 91.3 $\rightarrow$ \ 7.4 & 91.1 & 87.5 & 88.5 \\
    Pa $\rightarrow$ Cl & 86.4 $\rightarrow$ \ 9.5 & Re & 90.2 $\rightarrow$  16.1 & Sk & 77.6 $\rightarrow$  13.5 & 89.9 & 89.8 & 86.3 \\
    Pa $\rightarrow$ Re & 80.8 $\rightarrow$  23.1 & Cl & 85.0 $\rightarrow$  15.7 & Sk & 83.3 $\rightarrow$  13.8 & 93.3 & 92.5 & 81.9 \\
    Pa $\rightarrow$ Sk & 79.8 $\rightarrow$  10.3 & Cl & 85.8 $\rightarrow$  13.6 & Re & 90.7 $\rightarrow$  12.1 & 92.0 & 91.0 & 86.5 \\
    Re $\rightarrow$ Cl & 89.8 $\rightarrow$  12.6 & Pa & 79.4 $\rightarrow$  10.4 & Sk & 81.1 $\rightarrow$  20.4 & 93.4 & 95.2 & 84.2 \\
    Re $\rightarrow$ Pa & 89.1 $\rightarrow$  14.7 & Cl & 84.9 $\rightarrow$  13.7 & Sk & 75.9 $\rightarrow$ \ 8.9 & 90.9 & 93.0 & 86.0 \\
    Re $\rightarrow$ Sk & 92.0 $\rightarrow$  12.6 & Cl & 84.4 $\rightarrow$  19.7 & Pa & 79.5 $\rightarrow$ \ 8.7 & 93.5 & 97.8 & 84.4 \\
    Sk $\rightarrow$ Cl & 80.5 $\rightarrow$  13.8 & Pa & 80.2 $\rightarrow$  11.4 & Re & 88.0 $\rightarrow$  15.0 & 91.2 & 89.8 & 85.5 \\
    Sk $\rightarrow$ Pa & 80.9 $\rightarrow$  12.5 & Cl & 85.5 $\rightarrow$  11.2 & Re & 91.3 $\rightarrow$  15.0 & 93.0 & 92.9 & 86.0 \\
    Sk $\rightarrow$ Re & 80.9 $\rightarrow$  14.4 & Cl & 87.3 $\rightarrow$  19.6 & Pa & 80.1 $\rightarrow$  15.8 & 93.0 & 96.2 & 82.7 \\
    \bottomrule
  \end{tabular}}
  \label{ts_domain_detail}
\end{table}

\begin{table}[!t]
\renewcommand\arraystretch{1}
\setlength{\tabcolsep}{2pt}

  \centering
        \caption{The accuracy ($\%$) of target-specified AoD-IP on Office-Home-65~\cite{home}.}
  \resizebox{0.48\textwidth}{!}{
  \begin{tabular}{c|cccc|ccc}
    \toprule
    $X_a / X_u$ & Art & Clipart & Product & Real-World & $W_{u-a} \uparrow$ & $Drop_u \uparrow$ & $Drop_a \downarrow$ \\
    \midrule   
    Ar & 85.5 $\Rightarrow$ 85.4 & 68.0 $\Rightarrow$ 2.7 & 89.8 $\Rightarrow$ 4.2 & 88.5 $\Rightarrow$ 14.8 & 63.83 & 74.88 & 0.12 \\ 
    Cl & 81.0 $\Rightarrow$ 5.1 & 75.0 $\Rightarrow$ 74.9 & 90.8 $\Rightarrow$ 3.3 & 89.5 $\Rightarrow$ 6.2 & 61.57 & 82.24 & 0.07 \\ 
    Pr & 78.8 $\Rightarrow$ 4.8 & 73.3 $\Rightarrow$ 4.3 & 92.8 $\Rightarrow$ 92.6 & 87.5 $\Rightarrow$ 16.9 & 65.69 & 71.23 & 0.25 \\
    Re & 83.0 $\Rightarrow$ 16.8 & 71.3 $\Rightarrow$ 3.9 & 90.8 $\Rightarrow$ 14.7 & 90.0 $\Rightarrow$ 89.9 & 62.79 & 69.92 & 0.09 \\
    \hline
    Mean       & \multicolumn{4}{c|}{/} & 63.47 & 74.57 & 0.13 \\  
    \bottomrule
  \end{tabular}}
  \label{ts_home_AoD-IP}
\end{table}

\begin{table*}[!t]
\renewcommand\arraystretch{1}
\setlength{\tabcolsep}{1pt}
  \centering
      \caption{$W_{u-a}$, $Drop_u$, and $Drop_a$ of target-specified NTL, CUTI, CUPI, HNTL, SOPHON, IP-CLIP and AoD-IP. The best performance is indicated by the numbers in bold. Statistical significance (p-value $<$ 0.05~\cite{p1,p2}) is denoted with: $^{\ast}$(AoD-IP vs. others).}
  \resizebox{1\textwidth}{!}{
  \begin{tabular}{c|c|ccccccc|ccccccc|ccccccc}
    \toprule
    \multirow{3}{*}{Datasets} & \multirow{3}{*}{$X_a$} & \multicolumn{7}{c|}{$W_{u-a} \uparrow$} & \multicolumn{7}{c|}{$Drop_u \uparrow$} & \multicolumn{7}{c}{$Drop_a \downarrow$} \\
    \cline{3-23} 
    & & \multirow{2}{*}{\makecell[c]{NTL \\~\cite{NTL}}} & \multirow{2}{*}{\makecell[c]{CUTI \\~\cite{CUTI}}} & \multirow{2}{*}{\makecell[c]{CUPI \\~\cite{CUPI}}} & \multirow{2}{*}{\makecell[c]{HNTL \\~\cite{HNTL}}} & \multirow{2}{*}{\makecell[c]{SOPH \\~\cite{SOPHON}}} & \multirow{2}{*}{\makecell[c]{IPC \\~\cite{IPCLIP}}} & AoD & \multirow{2}{*}{\makecell[c]{NTL \\~\cite{NTL}}} & \multirow{2}{*}{\makecell[c]{CUTI \\~\cite{CUTI}}} & \multirow{2}{*}{\makecell[c]{CUPI \\~\cite{CUPI}}} & \multirow{2}{*}{\makecell[c]{HNTL \\~\cite{HNTL}}} &  \multirow{2}{*}{\makecell[c]{SOPH \\~\cite{SOPHON}}} & \multirow{2}{*}{\makecell[c]{IPC \\~\cite{IPCLIP}}} & AoD & \multirow{2}{*}{\makecell[c]{NTL \\~\cite{NTL}}} & \multirow{2}{*}{\makecell[c]{CUTI \\~\cite{CUTI}}} & \multirow{2}{*}{\makecell[c]{CUPI \\~\cite{CUPI}}} & \multirow{2}{*}{\makecell[c]{HNTL \\~\cite{HNTL}}} & \multirow{2}{*}{\makecell[c]{SOPH \\~\cite{SOPHON}}} & \multirow{2}{*}{\makecell[c]{IPC \\~\cite{IPCLIP}}} & AoD \\
    & & & & & & & & & & & & & & & \\
    \midrule 
    \multirow{4}{*}{\makecell[c]{Office-31 \\~\cite{office31}}} 
    & Am & 56.34 & 62.06 & 62.88 & 55.76 & 60.00 & 63.52 & {\bf 69.98} & 74.40 & 79.35 & 80.20 & 80.35 & 84.40 & 80.00 & {\bf 88.15} & 1.80 & 0.60 & 0.50 & 5.20 & 4.40 & {\bf 0.00} & {\bf 0.00} \\
    & Ds & 76.09 & 80.13 & 80.75 & 82.87 & 80.81 & 82.54 & {\bf 86.43} & 81.90 & 85.05 & 85.70 & 88.70 & 86.90 & 86.25 & {\bf 90.36} & 1.30 & 0.70 & 0.70 & 1.10 & 1.30 & {\bf 0.00} & {\bf 0.00} \\
    & We & 32.50 & 75.24 & 77.15 & 70.44 & 77.20 & 78.45 & {\bf 81.40} & 38.70 & 84.38 & 84.55 & 81.55 & 83.75 & 83.10 & {\bf 86.25} & 3.10 & 2.50 & 1.50 & 3.80 & 1.05 & {\bf 0.00} & {\bf 0.00} \\
\bottomrule
    \end{tabular}}
  \label{ts_compare}
\end{table*}

\begin{table*}[!t]
\renewcommand\arraystretch{1}
\setlength{\tabcolsep}{4pt}
  \centering
      \caption{The detailed results of authorization application AoD-IP on Office-Home-65~\cite{home}.}
  \resizebox{1\textwidth}{!}{
  \begin{tabular}{cc|cc|cc|cc|c}
    \toprule
	$X_a$ & $A^{ip}_a \rightarrow A^{ip}_u$ & $X_{e^1}$ & $A^{ip}_{e^1} \rightarrow A^{ip}_u$ & $X_{e^2}$ & $A^{ip}_{e^2} \rightarrow A^{ip}_u$ & $X_{e^3}$ & $A^{ip}_{e^3} \rightarrow A^{ip}_u$ & $R_a$ / $R_e$ / $R_u$ \\
    \midrule
	Ar$\dagger$ & 82.5 $\rightarrow$ 1.3 / 2.3 / \ 7.5 / \ 1.8 & Cl$\dagger$ & 56.5 $\rightarrow$ 2.0 / 4.0 / 8.0 / 7.5 & Pr$\dagger$ & 83.3 $\rightarrow$ 1.5 / 3.8 / \ 7.5 / 1.8 & Re$\dagger$ & 84.5 $\rightarrow$ 1.0 / 1.8 / 3.0 / 1.0 & \ 99.5 / \ 99.8 / 94.7  \\
	Cl$\dagger$ & 56.3 $\rightarrow$ 6.8 / 9.0 /  11.8 /  11.8 & Ar$\dagger$ & 81.3 $\rightarrow$ 2.5 / 1.5 / 8.0 / 4.8 & Pr$\dagger$ & 82.5 $\rightarrow$ 2.3 / 2.3 /  10.0 / 4.8 & Re$\dagger$ & 88.0 $\rightarrow$ 2.0 / 2.3 / 8.0 / 3.5 &  100.0 / \ 99.8 / 90.3  \\
	Pr$\dagger$ & 83.5 $\rightarrow$ 1.8 / 2.3 / \ 7.3 / \ 1.3 & Ar$\dagger$ & 82.3 $\rightarrow$ 0.8 / 1.0 / 2.8 / 1.3 & Cl$\dagger$ & 56.8 $\rightarrow$ 0.8 / 3.3 / \ 4.5 / 1.5 & Re$\dagger$ & 89.4 $\rightarrow$ 1.0 / 1.5 / 2.8 / 1.5 &  100.0 / \ 99.8 / 96.4  \\
	Re$\dagger$ & 88.0 $\rightarrow$ 2.8 / 4.8 /  10.3 / \ 3.3 & Ar$\dagger$ & 81.0 $\rightarrow$ 1.8 / 2.5 / 6.3 / 4.0 & Cl$\dagger$ & 56.5 $\rightarrow$ 1.8 / 3.5 / \ 5.5 / 3.0 & Pr$\dagger$ & 87.0 $\rightarrow$ 1.3 / 2.0 / 8.5 / 1.8 &  100.0 /  100.0 / 92.5  \\
    \bottomrule
  \end{tabular}}
  \label{aa_home_detail}
\end{table*}

\begin{table*}[!t]
\renewcommand\arraystretch{1}
\setlength{\tabcolsep}{1pt}
  \centering
      \caption{$D_{u-a}$, $A_u^{ip}$, and $A_a^{ip}$ of authorization application NTL, CUTI, CUPI, HNTL, SOPHON, IP-CLIP and AoD-IP.}
  \resizebox{1\textwidth}{!}{
  \begin{tabular}{c|c|ccccccc|ccccccc|ccccccc}
    \toprule
    \multirow{3}{*}{Datasets} & \multirow{3}{*}{$X_a$} & \multicolumn{7}{c|}{$D_{u-a} \uparrow$} & \multicolumn{7}{c|}{$A_u^{ip} \downarrow$} & \multicolumn{7}{c}{$A_a^{ip} \uparrow$} \\
    \cline{3-23} 
    & & \multirow{2}{*}{\makecell[c]{NTL \\~\cite{NTL}}} & \multirow{2}{*}{\makecell[c]{CUTI \\~\cite{CUTI}}} & \multirow{2}{*}{\makecell[c]{CUPI \\~\cite{CUPI}}} & \multirow{2}{*}{\makecell[c]{HNTL \\~\cite{HNTL}}} & \multirow{2}{*}{\makecell[c]{SOPH \\~\cite{SOPHON}}} & \multirow{2}{*}{\makecell[c]{IPC \\~\cite{IPCLIP}}} & AoD & \multirow{2}{*}{\makecell[c]{NTL \\~\cite{NTL}}} & \multirow{2}{*}{\makecell[c]{CUTI \\~\cite{CUTI}}} & \multirow{2}{*}{\makecell[c]{CUPI \\~\cite{CUPI}}} & \multirow{2}{*}{\makecell[c]{HNTL \\~\cite{HNTL}}} &  \multirow{2}{*}{\makecell[c]{SOPH \\~\cite{SOPHON}}} & \multirow{2}{*}{\makecell[c]{IPC \\~\cite{IPCLIP}}} & AoD & \multirow{2}{*}{\makecell[c]{NTL \\~\cite{NTL}}} & \multirow{2}{*}{\makecell[c]{CUTI \\~\cite{CUTI}}} & \multirow{2}{*}{\makecell[c]{CUPI \\~\cite{CUPI}}} & \multirow{2}{*}{\makecell[c]{HNTL \\~\cite{HNTL}}} & \multirow{2}{*}{\makecell[c]{SOPH \\~\cite{SOPHON}}} & \multirow{2}{*}{\makecell[c]{IPC \\~\cite{IPCLIP}}} & AoD \\
    & & & & & & & & & & & & & & & \\
    \midrule 
    \multirow{4}{*}{\makecell[c]{Office-31 \\~\cite{office31}}} 
    & Am$\dagger$ & 15.67 & 29.26 & 31.51 & 5.55 & 39.11 & 37.46 & \bf 44.64 & 37.43 & 20.83 & 16.92 & 5.73 & 9.57 & 3.53 & \bf 2.61 & 62.50 & 65.50 & 65.23 & 26.60 & 67.50 & 63.00 & \bf 68.13 \\
    & Ds$\dagger$ & 39.25 & 54.47 & 58.62 & 73.19 & 85.31 & 82.42 & \bf 87.55 & 50.50 & 36.53 & 33.30 & 21.37 & 10.00 & 9.77 & \bf 7.83 & 92.80 & 94.30 & 95.00 & 96.90 & 97.50 & 95.80 & \bf 97.57 \\
    & We$\dagger$ & 54.59 & 40.56 & 49.70 & \bf 78.36 & 74.09 & 56.45 & 75.96 & 21.30 & 30.60 & 25.68 & 18.77 & 12.40 & 15.53 & \bf 11.07 & 85.30 & 80.80 & 84.50 & \bf 98.40 & 92.50 & 83.30 & 92.87 \\
\bottomrule
    \end{tabular}}
  \label{aa_compare}
\end{table*}

\subsection{Target-Specified Model IP Protection}
In the most general scenario, where both the authorized and unauthorized domains are known and accessible, we evaluate the performance of AoD-IP under a controlled and reproducible task construction protocol. Specifically, for each dataset, one domain is randomly selected as the authorized domain, another as the unauthorized domain, and the remaining domains are designated as extended domains.
Following this strategy, we construct a total of 6, 12, and 12 tasks for Office-31, Office-Home-65, and Mini-DomainNet, respectively, as summarized in Tables~\ref{ts_office_detail},~\ref{ts_home_detail}, and~\ref{ts_domain_detail}.

The first column in each table details the training configuration (i.e., $X_a \rightarrow X_u$).
As observed, AoD-IP maintains strong performance on the authorized domain while exhibiting a substantial accuracy drop on the unauthorized domain.
The middle part of each table reports the model’s performance when switching to an extended domain after training (i.e., $X_{e^n} \rightarrow X_u$). AoD-IP continues to achieve stable predictions on the newly introduced extended domain while effectively suppressing its performance on the unauthorized domain.
The rightmost columns present the legality discrimination accuracies ($R_a$, $R_e$, and $R_u$), which measure the model’s ability to correctly identify samples from authorized and unauthorized domains, respectively.
The results range from 81.9\% to 100\%, with the majority exceeding 90\%, demonstrating the strong reliability of AoD-IP’s legality discrimination ability.

To provide a more intuitive comparison of model behavior with and without IP protection, we further present the results of AoD-IP and the supervised learning CLIP (SL-CLIP) baseline, as shown in Table~\ref{ts_home_AoD-IP}, \textit{Supplementary Table 7}, and \textit{Supplementary Table 20}.
In Table~\ref{ts_home_AoD-IP}, the vertical axis represents the authorized domains (including extended domains during testing), while the horizontal axis corresponds to the unauthorized domains.
Each cell reports a pair of values, where the left side of "$\Rightarrow$" denotes the accuracy of the SL-CLIP on the unauthorized domain ($A^{sl}_u$), and the right side denotes the accuracy of AoD-IP ($A^{ip}_u$), averaged over all corresponding tasks listed in Table~\ref{ts_home_detail}.
The results clearly indicate that, without IP protection, models can be easily transferred to unauthorized domains and still achieve high prediction accuracy.
In contrast, AoD-IP exhibits a substantial performance drop on unauthorized domains, effectively preventing illegal transfer, with an average $Drop_u$ of 74.57\%.
Meanwhile, AoD-IP imposes only a marginal negative effect on the authorized domain, achieving a minimal $Drop_a$ of 0.13\% and a strong overall trade-off with $W_{u-a}$ = 63.47\%.

Finally, Table~\ref{ts_compare} compares AoD-IP with other IP protection methods.
Across nearly all tasks, AoD-IP consistently achieves the best comprehensive performance. Although IP-CLIP occasionally attains comparable $Drop_a$, it suffers from relatively weaker $Drop_u$, leading to a slightly lower overall metric $W_{u-a}$. In contrast, HNTL sometimes achieves favorable $Drop_u$, but at the cost of severely degrading performance on the authorized domain, where $Drop_a$ reaches up to 28.83\%, which is unacceptable in practical applications. Overall, AoD-IP demonstrates superior flexible domain-switching flexibility and robust protection capability, highlighting its potential for secure and flexible deployment in dynamic real-world environments.

Detailed results of all compared methods are provided in the \textit{Supplementary Tables 1–6} and \textit{8–19}.

\subsection{Authorization Application Model IP Protection}

In a more realistic and challenging scenario, the objective is twofold: to prevent the model from being transferred to unknown domains, and to ensure that only authorized users can access and utilize it. This configuration better mirrors real-world deployment conditions and is referred to as applicability authorization. Specifically, for each dataset, one domain is randomly selected and embedded with a private watermark, designating it as the authorized domain, while the remaining domains are treated as unauthorized during evaluation. 
Under this setting, we construct 3, 4, and 4 applicability authorization tasks on Office-31~\cite{office31}, Office-Home-65~\cite{home}, and Mini-DomainNet~\cite{mini}, respectively, as summarized in Table~\ref{aa_home_detail} and \textit{Supplementary Tables~21–22}.

In each table, the vertical axis lists the authorized domain ($X_a$) and extended domains ($X_{e^n}$), with “$\dagger$” marking watermarked data, while the horizontal axis lists unauthorized domains.
AoD-IP performs strongly only on the authorized or extended domains, but its accuracy drops sharply once the watermark is removed or when applied to unauthorized domains, demonstrating both flexible domain switching and strong IP protection.

Table~\ref{aa_compare} presents the comparative results between AoD-IP and other methods. Even under more complex settings, AoD-IP consistently achieves superior performance, with the overall metric $D_{u-a}$ reaching 69.39\%, 58.64\%, and 56.59\% on the respective benchmarks.
Notably, although HNTL attains lower $A_u^{ip}$, its performance on the authorized domain sometimes collapses to near-random levels, leading to an inferior overall score.
In addition, AoD-IP achieves an average legality discrimination accuracy exceeding 97\%, demonstrating its superior flexibility and robustness for IP protection in dynamic environments.

\section{Conclusion}

The rapid advancement of VLMs has raised growing concerns over model intellectual property (IP) security. We propose (\textbf{AoD-IP}), a novel dynamic authorization framework that introduces a lightweight authorization module for \textbf{flexible, user-controlled authorization} and a dual-path inference strategy for simultaneous legality verification and task prediction. Experiments demonstrate that AoD-IP delivers robust legality discrimination and flexible deployment capabilities. Future work will extend AoD-IP to broader tasks (e.g., VQA and image generation) and explore more diverse datasets and architectures to further enhance its generalizability in realistic environments.
\section*{Acknowledgments}
This work is supported by the National Natural Science Foundation of China (Nos. 62136004, 62276130), the Key Research and Development Plan of Jiangsu Province (No. BE2022842), and H. Fu’s A*STAR Central Research Fund.
{
    \small
    \bibliographystyle{ieeenat_fullname}
    \bibliography{main}
}


\clearpage
\setcounter{page}{1}

\begin{figure*}[!t]
  \centering
   \includegraphics[width=1\linewidth,trim=60 670 60 90,clip]{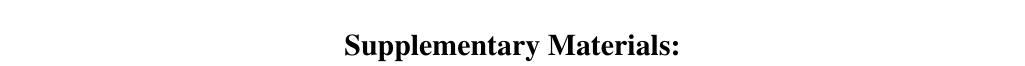}
   \label{figure2} 
\end{figure*}

\setcounter{table}{0}

\begin{table*}[!t]
\renewcommand\arraystretch{1.2}
  \centering
      \caption{The accuracy ($\%$) of target-specified NTL on Office-31. The vertical/horizontal axis denotes the authorized/unauthorized domain. In each task, the left of '\(\Rightarrow\)' shows the test accuracy of SL-CLIP on the unauthorized domain, while the right side presents the accuracy of NTL.}
  \resizebox{1\textwidth}{!}{
\begin{tabular}{c|ccc|ccc}
    \toprule
    Authorized/Unauthorized & Amazon & Dslr & Webcam & $W_{u-a} \uparrow$ & $Drop_u \uparrow$ & $Drop_a \downarrow$ \\
    \midrule   
    Amazon & 79.4 $\Rightarrow$   77.6 & 87.5 $\Rightarrow$ \ 10.0 & 88.8  $\Rightarrow$ \ 17.5 & 56.34 & 74.40 & 1.80 \\ 
    Dslr   & 83.8 $\Rightarrow$ \ 10.0 & 95.7 $\Rightarrow$  94.4 & 98.8 $\Rightarrow$ \ 8.8 & 76.09 & 81.90 & 1.30 \\ 
    Webcam & 80.0 $\Rightarrow$ \ 3.8 & 92.5 $\Rightarrow$ \ 91.3 & 94.4  $\Rightarrow$  91.3 & 32.50 & 38.70 & 3.10 \\
    \midrule
    Mean   & \multicolumn{3}{c|}{/} & 54.98 & 65.00 & 2.07 \\
    \bottomrule
    \end{tabular}}
  \label{ts_office31_ntl}
\end{table*}

\begin{table*}[!t]
\renewcommand\arraystretch{1.2}
  \centering
      \caption{The accuracy ($\%$) of target-specified CUTI-Domain on Office-31. The vertical/horizontal axis denotes the authorized/unauthorized domain. In each task, the left of '\(\Rightarrow\)' shows the test accuracy of SL-CLIP on the unauthorized domain, while the right side presents the accuracy of CUTI-Domain.}
  \resizebox{1\textwidth}{!}{
  \begin{tabular}{c|ccc|ccc}
    \toprule
    Authorized/Unauthorized & Amazon & Dslr & Webcam & $W_{u-a} \uparrow$ & $Drop_u \uparrow$ & $Drop_a \downarrow$ \\
    \midrule   
    Amazon & 79.4 $\Rightarrow$  78.8 & 87.5 $\Rightarrow$ \ 6.3 & 88.8  $\Rightarrow$ \ 11.3 & 62.06 & 79.35 & 0.60 \\ 
    Dslr   & 83.8 $\Rightarrow$ \ 5.0 & 95.7 $\Rightarrow$ 95.0 & 98.8 $\Rightarrow$ \ 7.5 & 80.13 & 85.05 & 0.70 \\ 
    Webcam & 80.0 $\Rightarrow$ \ 1.3 & 92.5 $\Rightarrow$ \ 2.5 & 94.4  $\Rightarrow$  91.9 & 75.24 & 84.38 & 2.50 \\
    \midrule
    Mean   & \multicolumn{3}{c|}{/} & 72.48 & 82.93 & 1.27 \\
    \bottomrule
    \end{tabular}}
  \label{ts_office31_cuti}
\end{table*}

\begin{table*}[!t]
\renewcommand\arraystretch{1.2}
  \centering
      \caption{The accuracy ($\%$) of target-specified CUPI-Domain on Office-31. The vertical/horizontal axis denotes the authorized/unauthorized domain. In each task, the left of '\(\Rightarrow\)' shows the test accuracy of SL-CLIP on the unauthorized domain, while the right side presents the accuracy of CUPI-Domain.}
  \resizebox{1\textwidth}{!}{
  \begin{tabular}{c|ccc|ccc}
    \toprule
    Authorized/Unauthorized & Amazon & Dslr & Webcam & $W_{u-a} \uparrow$ & $Drop_u \uparrow$ & $Drop_a \downarrow$ \\
    \midrule   
    Amazon & 79.4 $\Rightarrow$ 78.9 & 87.5 $\Rightarrow$ 5.2 & 88.8 $\Rightarrow$ 10.7 & 62.88 & 80.20 & 0.50 \\ 
    Dslr   & 83.8 $\Rightarrow$ 4.3 & 95.7 $\Rightarrow$ 95.0 & 98.8 $\Rightarrow$ 6.9 & 80.75 & 85.70 & 0.70 \\ 
    Webcam & 80.0 $\Rightarrow$ 0.9 & 92.5 $\Rightarrow$ 2.5 & 94.4 $\Rightarrow$ 92.9 & 77.15 & 84.55 & 1.50 \\
    \midrule
    Mean   & \multicolumn{3}{c|}{/} & 73.60 & 83.48 & 0.90 \\
    \bottomrule
    \end{tabular}}
  \label{ts_office31_cupi}
\end{table*}

\begin{table*}[!t]
\renewcommand\arraystretch{1.2}
  \centering
      \caption{The accuracy ($\%$) of target-specified HNTL on Office-31. The vertical/horizontal axis denotes the authorized/unauthorized domain. In each task, the left of '\(\Rightarrow\)' shows the test accuracy of SL-CLIP on the unauthorized domain, while the right side presents the accuracy of HNTL.}
  \resizebox{1\textwidth}{!}{
  \begin{tabular}{c|ccc|ccc}
    \toprule
    Authorized/Unauthorized & Amazon & Dslr & Webcam & $W_{u-a} \uparrow$ & $Drop_u \uparrow$ & $Drop_a \downarrow$ \\
    \midrule   
    Amazon & 79.4 $\Rightarrow$  74.2 & 87.5 $\Rightarrow$ 12.5 & 88.8  $\Rightarrow$  3.1 & 55.76 & 80.35 & 5.20 \\ 
    Dslr   & 83.8 $\Rightarrow$  3.1 & 95.7 $\Rightarrow$ 94.6 & 98.8 $\Rightarrow$  2.1 & 82.87 & 88.70 & 1.10 \\ 
    Webcam & 80.0 $\Rightarrow$  6.3 & 92.5 $\Rightarrow$  3.1 & 94.4  $\Rightarrow$ 90.6 & 70.44 & 81.55 & 3.80 \\
    \midrule
    Mean   & \multicolumn{3}{c|}{/} & 69.69 & 83.53 & 3.37 \\
    \bottomrule
    \end{tabular}}
  \label{ts_office31_hntl}
\end{table*}

\begin{table*}[!t]
\renewcommand\arraystretch{1.2}
  \centering
      \caption{The accuracy ($\%$) of target-specified SOPHON on Office-31. The vertical/horizontal axis denotes the authorized/unauthorized domain. In each task, the left of '\(\Rightarrow\)' shows the test accuracy of SL-CLIP on the unauthorized domain, while the right side presents the accuracy of SOPHON.}
  \resizebox{1\textwidth}{!}{
  \begin{tabular}{c|ccc|ccc}
    \toprule
    Authorized/Unauthorized & Amazon & Dslr & Webcam & $W_{u-a} \uparrow$ & $Drop_u \uparrow$ & $Drop_a \downarrow$ \\
    \midrule   
    Amazon & 79.4 $\Rightarrow$ 75.0 & 87.5 $\Rightarrow$ 2.5 & 88.8 $\Rightarrow$ 5.0 & 60.00 & 84.40 & 4.40 \\ 
    Dslr   & 83.8 $\Rightarrow$ 3.8 & 95.7 $\Rightarrow$ 94.4 & 98.8 $\Rightarrow$ 5.0 & 80.81 & 86.90 & 1.30 \\ 
    Webcam & 80.0 $\Rightarrow$ 2.5 & 92.5 $\Rightarrow$ 2.5 & 94.4 $\Rightarrow$ 93.4 & 77.20 & 83.75 & 1.05 \\
    \midrule
    Mean   & \multicolumn{3}{c|}{/} & 72.67 & 85.02 & 2.25 \\
    \bottomrule
  \end{tabular}}
  \label{ts_office31_sophon}
\end{table*}

\begin{table*}[!t]
\renewcommand\arraystretch{1.2}
  \centering
      \caption{The accuracy ($\%$) of target-specified IP-CLIP on Office-31. The vertical/horizontal axis denotes the authorized/unauthorized domain. In each task, the left of '\(\Rightarrow\)' shows the test accuracy of SL-CLIP on the unauthorized domain, while the right side presents the accuracy of IP-CLIP.}
  \resizebox{1\textwidth}{!}{
  \begin{tabular}{c|ccc|ccc}
    \toprule
    Authorized/Unauthorized & Amazon & Dslr & Webcam & $W_{u-a} \uparrow$ & $Drop_u \uparrow$ & $Drop_a \downarrow$ \\
    \midrule   
    Amazon & 79.4 $\Rightarrow$  79.4 & 87.5 $\Rightarrow$ \ 7.5 & 88.8 $\Rightarrow$ \ 8.8 & 63.52 & 80.00 & 0.00 \\ 
    Dslr   & 83.8 $\Rightarrow$ \ 3.8 & 95.7 $\Rightarrow$  95.7 & 98.8 $\Rightarrow$ \ 6.3 & 82.54 & 86.25 & 0.00 \\ 
    Webcam & 80.0 $\Rightarrow$ \ 3.8 & 92.5 $\Rightarrow$ \ 2.5 & 94.4 $\Rightarrow$  94.4 & 78.45 & 83.10 & 0.00 \\
    \midrule
    Mean   & \multicolumn{3}{c|}{/} & 74.84 & 83.12 & 0.00 \\
    \bottomrule
  \end{tabular}}
  \label{ts_office31_ipclip}
\end{table*}

\begin{table*}[!t]
\renewcommand\arraystretch{1.2}
  \centering
      \caption{The accuracy ($\%$) of target-specified AoD-IP on Office-31. The vertical/horizontal axis denotes the authorized/unauthorized domain. In each task, the left of '\(\Rightarrow\)' shows the test accuracy of SL-CLIP on the unauthorized domain, while the right side presents the accuracy of AoD-IP.}
  \resizebox{1\textwidth}{!}{
  \begin{tabular}{c|ccc|ccc}
    \toprule
    Authorized/Unauthorized & Amazon & Dslr & Webcam & $W_{u-a} \uparrow$ & $Drop_u \uparrow$ & $Drop_a \downarrow$ \\
    \midrule   
    Amazon & 79.4 $\Rightarrow$ 79.4 & 87.5 $\Rightarrow$ 0.0 & 88.8 $\Rightarrow$ 0.0 & 69.98 & 88.15 & 0.00 \\ 
    Dslr   & 83.8 $\Rightarrow$ 0.6 & 95.7 $\Rightarrow$ 95.7 & 98.8 $\Rightarrow$ 1.3 & 86.43 & 90.36 & 0.00 \\ 
    Webcam & 80.0 $\Rightarrow$ 0.0 & 92.5 $\Rightarrow$ 0.0 & 94.4 $\Rightarrow$ 94.4 & 81.40 & 86.25 & 0.00 \\
    \midrule
    Mean   & \multicolumn{3}{c|}{/} & 79.27 & 88.25 & 0.00 \\
    \bottomrule
  \end{tabular}}
  \label{ts_office31_AoD-IP}
\end{table*}

\begin{table*}[!t]
\renewcommand\arraystretch{1.2}
  \centering
        \caption{The accuracy ($\%$) of target-specified NTL on Office-Home-65. The vertical/horizontal axis denotes the authorized/unauthorized domain. In each task, the left of '\(\Rightarrow\)' shows the test accuracy of SL-CLIP on the unauthorized domain, while the right side presents the accuracy of NTL.}
  \resizebox{1\textwidth}{!}{
  \begin{tabular}{c|cccc|ccc}
    \toprule
    Authorized/Unauthorized & Art & Clipart & Product & Real-World & $W_{u-a} \uparrow$ & $Drop_u \uparrow$ & $Drop_a \downarrow$ \\
    \midrule   
    Art        & 85.5 $\Rightarrow$ 85.4 & 68.0 $\Rightarrow$  23.8 & 89.8 $\Rightarrow$ 86.5 & 88.5 $\Rightarrow$ 88.5 & 13.44 & 15.83 & 0.10 \\ 
    Clipart    & 81.0 $\Rightarrow$ 18.3 & 75.0 $\Rightarrow$  74.7 & 90.8 $\Rightarrow$ \ 15.0 & 89.5 $\Rightarrow$ 31.0 & 48.83 & 65.67 & 0.30  \\ 
    Product    & 78.8 $\Rightarrow$ 11.0 & 73.3 $\Rightarrow$ \ 13.8 & 92.8 $\Rightarrow$  92.8 & 87.5 $\Rightarrow$ 85.8 & 39.90 & 43.00 & 0.00 \\
    Real-World  & 83.0 $\Rightarrow$ 80.8 & 71.3 $\Rightarrow$ \ 7.8 & 90.8 $\Rightarrow$  52.5 & 90.0 $\Rightarrow$ 88.1 & 28.87 & 34.67 & 1.90 \\
    \hline
    Mean       & \multicolumn{4}{c|}{/} &32.76 & 39.79 & 0.57 \\  
    \bottomrule
  \end{tabular}}
  \label{ts_home_ntl}
\end{table*}

\begin{table*}[!t]
\renewcommand\arraystretch{1.2}
  \centering
        \caption{The accuracy ($\%$) of target-specified CUTI-Domain on Office-Home-65. The vertical/horizontal axis denotes the authorized/unauthorized domain. In each task, the left of '\(\Rightarrow\)' shows the test accuracy of SL-CLIP on the unauthorized domain, while the right side presents the accuracy of CUTI-Domain.}
  \resizebox{1\textwidth}{!}{
  \begin{tabular}{c|cccc|ccc}
    \toprule
    Authorized/Unauthorized & Art & Clipart & Product & Real-World & $W_{u-a} \uparrow$ & $Drop_u \uparrow$ & $Drop_a \downarrow$ \\
    \midrule   
    Art        & 85.5 $\Rightarrow$ 82.5 & 68.0 $\Rightarrow$  16.8 & 89.8 $\Rightarrow$ 21.3 & 88.5 $\Rightarrow$ 48.0 & 41.58 & 53.40 & 3.00 \\ 
    Clipart    & 81.0 $\Rightarrow$ 15.8 & 75.0 $\Rightarrow$  75.0 & 90.8 $\Rightarrow$ \ 9.0 & 89.5 $\Rightarrow$ 19.3 & 53.37 & 72.40 & 0.63  \\ 
    Product    & 78.8 $\Rightarrow$ 16.3 & 73.3 $\Rightarrow$ \ 10.8 & 92.8 $\Rightarrow$  92.4 & 87.5 $\Rightarrow$ 27.0 & 56.82 & 61.83 & 0.37 \\
    Real-World  & 83.0 $\Rightarrow$ 30.3 & 71.3 $\Rightarrow$ \ 10.0 & 90.8 $\Rightarrow$  32.8 & 90.0 $\Rightarrow$ 88.5 & 49.41 & 57.33 & 1.50 \\
    \hline
    Mean       & \multicolumn{4}{c|}{/} & 50.29 & 61.24 & 1.38 \\  
    \bottomrule
  \end{tabular}}
  \label{ts_home_cuti}
\end{table*}

\begin{table*}[!t]
\renewcommand\arraystretch{1.2}
  \centering
        \caption{The accuracy ($\%$) of target-specified CUPI-Domain on Office-Home-65. The vertical/horizontal axis denotes the authorized/unauthorized domain. In each task, the left of '\(\Rightarrow\)' shows the test accuracy of SL-CLIP on the unauthorized domain, while the right side presents the accuracy of CUPI-Domain.}
  \resizebox{1\textwidth}{!}{
  \begin{tabular}{c|cccc|ccc}
    \toprule
    Authorized/Unauthorized & Art & Clipart & Product & Real-World & $W_{u-a} \uparrow$ & $Drop_u \uparrow$ & $Drop_a \downarrow$ \\
    \midrule   
    Art        & 85.5 $\Rightarrow$ 83.4 & 68.0 $\Rightarrow$ 13.0 & 89.8 $\Rightarrow$ 21.0 & 88.5 $\Rightarrow$ 47.3 & 44.12 & 55.00 & 2.10 \\ 
    Clipart    & 81.0 $\Rightarrow$ 15.0 & 75.0 $\Rightarrow$ 74.9 & 90.8 $\Rightarrow$ 10.0 & 89.5 $\Rightarrow$ 16.1 & 54.90 & 73.40 & 0.10 \\ 
    Product    & 78.8 $\Rightarrow$ 15.9 & 73.3 $\Rightarrow$ 9.8 & 92.8 $\Rightarrow$ 92.5 & 87.5 $\Rightarrow$ 25.1 & 57.94 & 62.93 & 0.30 \\
    Real-World  & 83.0 $\Rightarrow$ 25.4 & 71.3 $\Rightarrow$ 10.0 & 90.8 $\Rightarrow$ 26.7 & 90.0 $\Rightarrow$ 89.5 & 54.15 & 61.00 & 0.50 \\
    \midrule
    Mean       & \multicolumn{4}{c|}{/} & 52.78 & 63.08 & 0.75 \\  
    \bottomrule
  \end{tabular}}
  \label{ts_home_cupi}
\end{table*}

\begin{table*}[!t]
\renewcommand\arraystretch{1.2}
  \centering
        \caption{The accuracy ($\%$) of target-specified HNTL on Office-Home-65. The vertical/horizontal axis denotes the authorized/unauthorized domain. In each task, the left of '\(\Rightarrow\)' shows the test accuracy of SL-CLIP on the unauthorized domain, while the right side presents the accuracy of HNTL.}
  \resizebox{1\textwidth}{!}{
  \begin{tabular}{c|cccc|ccc}
    \toprule
    Authorized/Unauthorized & Art & Clipart & Product & Real-World & $W_{u-a} \uparrow$ & $Drop_u \uparrow$ & $Drop_a \downarrow$ \\
    \midrule   
    Art        & 85.5 $\Rightarrow$ 70.1 & 68.0 $\Rightarrow$ 4.1 & 89.8 $\Rightarrow$ 7.0 & 88.5 $\Rightarrow$ 22.9 & 38.81 & 70.77 & 15.40 \\ 
    Clipart    & 81.0 $\Rightarrow$ 5.5 & 75.0 $\Rightarrow$ 46.2 & 90.8 $\Rightarrow$ 5.2 & 89.5 $\Rightarrow$ 9.1 & 23.85 & 80.50 & 28.83 \\ 
    Product    & 78.8 $\Rightarrow$ 4.7 & 73.3 $\Rightarrow$ 5.7 & 92.8 $\Rightarrow$ 70.7 & 87.5 $\Rightarrow$ 27.3 & 31.96 & 67.30 & 22.10 \\
    Real-World  & 83.0 $\Rightarrow$ 35.9 & 71.3 $\Rightarrow$ 6.2 & 90.8 $\Rightarrow$ 26.6 & 90.0 $\Rightarrow$ 78.8 & 37.51 & 58.80 & 11.20 \\
    \hline
    Mean       & \multicolumn{4}{c|}{/} & 33.03 & 69.34 & 19.38 \\  
    \bottomrule
  \end{tabular}}
  \label{ts_home_hntl}
\end{table*}

\begin{table*}[!t]
\renewcommand\arraystretch{1.2}
  \centering
        \caption{The accuracy ($\%$) of target-specified SOPHON on Office-Home-65. The vertical/horizontal axis denotes the authorized/unauthorized domain. In each task, the left of '\(\Rightarrow\)' shows the test accuracy of SL-CLIP on the unauthorized domain, while the right side presents the accuracy of SOPHON.}
  \resizebox{1\textwidth}{!}{
  \begin{tabular}{c|cccc|ccc}
    \toprule
    Authorized/Unauthorized & Art & Clipart & Product & Real-World & $W_{u-a} \uparrow$ & $Drop_u \uparrow$ & $Drop_a \downarrow$ \\
    \midrule   
    Art        & 85.5 $\Rightarrow$ 73.1 & 68.0 $\Rightarrow$ 21.5 & 89.8 $\Rightarrow$ 45.8 & 88.5 $\Rightarrow$ 64.0 & 18.99 & 38.33 & 12.37 \\ 
    Clipart    & 81.0 $\Rightarrow$ 13.5 & 75.0 $\Rightarrow$ 74.5 & 90.8 $\Rightarrow$ 21.2 & 89.5 $\Rightarrow$ 22.0 & 50.44 & 68.20 & 0.50 \\ 
    Product    & 78.8 $\Rightarrow$ 29.2 & 73.3 $\Rightarrow$ 18.2 & 92.8 $\Rightarrow$ 89.0 & 87.5 $\Rightarrow$ 58.5 & 36.28 & 44.57 & 3.80 \\
    Real-World  & 83.0 $\Rightarrow$ 51.5 & 71.3 $\Rightarrow$ 24.0 & 90.8 $\Rightarrow$ 61.3 & 90.0 $\Rightarrow$ 80.5 & 21.38 & 36.10 & 9.53 \\
    \hline
    Mean       & \multicolumn{4}{c|}{/} & 31.77 & 46.80 & 6.55 \\  
    \bottomrule
  \end{tabular}}
  \label{ts_home_sophon}
\end{table*}

\begin{table*}[!t]
\renewcommand\arraystretch{1.2}
  \centering
        \caption{The accuracy ($\%$) of target-specified IP-CLIP on Office-Home-65. The vertical/horizontal axis denotes the authorized/unauthorized domain. In each task, the left of '\(\Rightarrow\)' shows the test accuracy of SL-CLIP on the unauthorized domain, while the right side presents the accuracy of IP-CLIP.}
  \resizebox{1\textwidth}{!}{
  \begin{tabular}{c|cccc|ccc}
    \toprule
    Authorized/Unauthorized & Art & Clipart & Product & Real-World & $W_{u-a} \uparrow$ & $Drop_u \uparrow$ & $Drop_a \downarrow$ \\
    \midrule   
    Art        & 85.5 $\Rightarrow$ 85.2 & 68.0 $\Rightarrow$  12.8 & 89.8 $\Rightarrow$  15.0 & 88.5 $\Rightarrow$ 34.5 & 52.00 & 61.33 & 0.30 \\ 
    Clipart    & 81.0 $\Rightarrow$ 11.8 & 75.0 $\Rightarrow$  74.9 & 90.8 $\Rightarrow$ \ 5.3 & 89.5 $\Rightarrow$ 17.8 & 56.45 & 75.47 & 0.10 \\ 
    Product    & 78.8 $\Rightarrow$ 14.0 & 73.3 $\Rightarrow$ \ 8.5 & 92.8 $\Rightarrow$  92.5 & 87.5 $\Rightarrow$ 25.8 & 58.71 & 63.77 & 0.30 \\
    Real-World  & 83.0 $\Rightarrow$ 30.3 & 71.3 $\Rightarrow$ \ 7.5 & 90.8 $\Rightarrow$  29.3 & 90.0 $\Rightarrow$ 89.9 & 53.25 & 59.33 & 0.10 \\
    \hline
    Mean       & \multicolumn{4}{c|}{/} & 55.10 & 64.98 & 0.20 \\  
    \bottomrule
  \end{tabular}}
  \label{ts_home_ipclip}
\end{table*}

\begin{table*}[!t]
\renewcommand\arraystretch{1.2}
  \centering
      \caption{The accuracy ($\%$) of target-specified NTL on Mini-DomainNet. The vertical/horizontal axis denotes the authorized/unauthorized domain. In each task, the left of '\(\Rightarrow\)' shows the test accuracy of SL-CLIP on the unauthorized domain, while the right side presents the accuracy of NTL.}
  \resizebox{1\textwidth}{!}{
  \begin{tabular}{c|cccc|cccc}
    \toprule
    Authorized/Unauthorized & Clipart & Painting & Real & Sketch & $W_{u-a} \uparrow$ & $Drop_u \uparrow$ & $Drop_a \downarrow$ \\
    \midrule   
    Clipart   & 85.1 $\Rightarrow$ 84.5  & 79.8  $\Rightarrow$ 33.3 & 89.8  $\Rightarrow$ 34.0 & 78.7  $\Rightarrow$ 42.1 & 38.62 & 46.30 & 0.60 \\
    Painting  & 83.8 $\Rightarrow$ 11.9  & 81.4  $\Rightarrow$ 79.8 & 89.1  $\Rightarrow$ 60.5 & 78.4  $\Rightarrow$ 17.5 & 41.66 & 53.80 & 1.60 \\
    Real      & 84.6 $\Rightarrow$ 26.4  & 80.5  $\Rightarrow$ 30.8 & 90.6  $\Rightarrow$ 89.8 & 80.0  $\Rightarrow$ 10.8 & 52.29 & 59.03 & 0.80 \\
    Sketch    & 84.3 $\Rightarrow$ 83.0  & 79.1  $\Rightarrow$ 32.7 & 90.3  $\Rightarrow$ 9.7  & 80.7  $\Rightarrow$ 80.1 & 33.78 & 42.77 & 0.60 \\
    \hline
    Mean      & \multicolumn{4}{c|}{/} & 41.59 & 50.48 & 1.00  \\  
    \bottomrule
  \end{tabular}}
  \label{ts_mini_ntl}
\end{table*}

\begin{table*}[!t]
\renewcommand\arraystretch{1.2}
  \centering
      \caption{The accuracy ($\%$) of target-specified CUTI-Domain on Mini-DomainNet. The vertical/horizontal axis denotes the authorized/unauthorized domain. In each task, the left of '\(\Rightarrow\)' shows the test accuracy of SL-CLIP on the unauthorized domain, while the right side presents the accuracy of CUTI-Domain.}
  \resizebox{1\textwidth}{!}{
  \begin{tabular}{c|cccc|cccc}
    \toprule
    Authorized/Unauthorized & Clipart & Painting & Real & Sketch & $W_{u-a} \uparrow$ & $Drop_u \uparrow$ & $Drop_a \downarrow$ \\
    \midrule   
        Clipart   & 85.1  $\Rightarrow$ 84.9  & 79.8  $\Rightarrow$ 22.7  & 89.8  $\Rightarrow$ 23.7  & 78.7  $\Rightarrow$ 23.7  & 50.26 & 59.40 & 0.20 \\
        Painting  & 83.8  $\Rightarrow$ 17.9  & 81.4  $\Rightarrow$ 76.1  & 89.1  $\Rightarrow$ 15.6  & 78.4  $\Rightarrow$ 17.1  & 46.88 & 66.90 & 5.30 \\
        Real      & 84.6  $\Rightarrow$ 24.9  & 80.5  $\Rightarrow$ 23.8  & 90.6  $\Rightarrow$ 89.5  & 80.0  $\Rightarrow$ 9.5  & 54.77 & 62.30 & 1.10 \\
        Sketch    & 84.3  $\Rightarrow$ 24.0  & 79.1  $\Rightarrow$ 25.2  & 90.3  $\Rightarrow$ 10.8  & 80.7  $\Rightarrow$ 80.0  & 51.09 & 64.57 & 0.70 \\ 
    \hline
    Mean      & \multicolumn{4}{c|}{/} & 50.75 & 63.29 & 2.20 \\  
    \bottomrule
  \end{tabular}}
  \label{ts_mini_cuti}
\end{table*}

\begin{table*}[!t]
\renewcommand\arraystretch{1.2}
  \centering
      \caption{The accuracy ($\%$) of target-specified CUPI-Domain on Mini-DomainNet. The vertical/horizontal axis denotes the authorized/unauthorized domain. In each task, the left of '\(\Rightarrow\)' shows the test accuracy of SL-CLIP on the unauthorized domain, while the right side presents the accuracy of CUPI-Domain..}
  \resizebox{1\textwidth}{!}{
  \begin{tabular}{c|cccc|cccc}
    \toprule
    Authorized/Unauthorized & Clipart & Painting & Real & Sketch & $W_{u-a} \uparrow$ & $Drop_u \uparrow$ & $Drop_a \downarrow$ \\
    \midrule   
        Clipart   & 85.1 $\Rightarrow$ 84.9 & 79.8 $\Rightarrow$ 20.8 & 89.8 $\Rightarrow$ 21.4 & 78.7 $\Rightarrow$ 19.9 & 52.52 & 62.07 & 0.20 \\
        Painting  & 83.8 $\Rightarrow$ 16.7 & 81.4 $\Rightarrow$ 78.3 & 89.1 $\Rightarrow$ 13.7 & 78.4 $\Rightarrow$ 16.9 & 50.82 & 68.00 & 3.10 \\
        Real      & 84.6 $\Rightarrow$ 25.1 & 80.5 $\Rightarrow$ 20.7 & 90.6 $\Rightarrow$ 89.7 & 80.0 $\Rightarrow$ 8.9 & 56.12 & 63.47 & 0.90 \\
        Sketch    & 84.3 $\Rightarrow$ 22.4 & 79.1 $\Rightarrow$ 23.7 & 90.3 $\Rightarrow$ 9.4 & 80.7 $\Rightarrow$ 80.3 & 52.73 & 66.07 & 0.40 \\ 
    \hline
    Mean      & \multicolumn{4}{c|}{/} & 53.05 & 64.90 & 1.40 \\  
    \bottomrule
  \end{tabular}}
  \label{ts_mini_cupi}
\end{table*}

\begin{table*}[!t]
\renewcommand\arraystretch{1.2}
  \centering
      \caption{The accuracy ($\%$) of target-specified HNTL on Mini-DomainNet. The vertical/horizontal axis denotes the authorized/unauthorized domain. In each task, the left of '\(\Rightarrow\)' shows the test accuracy of SL-CLIP on the unauthorized domain, while the right side presents the accuracy of HNTL.}
  \resizebox{1\textwidth}{!}{
  \begin{tabular}{c|cccc|cccc}
    \toprule
    Authorized/Unauthorized & Clipart & Painting & Real & Sketch & $W_{u-a} \uparrow$ & $Drop_u \uparrow$ & $Drop_a \downarrow$ \\
    \midrule   
    Clipart   & 85.1 $\Rightarrow$ 59.6 & 79.8 $\Rightarrow$ 7.2 & 89.8 $\Rightarrow$ 11.2 & 78.7 $\Rightarrow$ 5.3 & 29.42 & 74.87 & 25.50 \\ 
    Painting  & 83.8 $\Rightarrow$ 6.1 & 81.4 $\Rightarrow$ 63.3 & 89.1 $\Rightarrow$ 18.3 & 78.4 $\Rightarrow$ 3.6 & 35.66 & 74.43 & 18.10 \\ 
    Real      & 84.6 $\Rightarrow$ 12.8 & 80.5 $\Rightarrow$ 19.9 & 90.6 $\Rightarrow$ 84.1 & 80.0 $\Rightarrow$ 6.6 & 52.23 & 68.60 & 6.50 \\ 
    Sketch    & 84.3 $\Rightarrow$ 4.9 & 79.1 $\Rightarrow$ 3.6 & 90.3 $\Rightarrow$ 6.3 & 80.7 $\Rightarrow$ 42.0 & 17.19 & 79.63 & 38.70 \\ 
    \hline
    Mean      & \multicolumn{4}{c|}{/} & 33.62 & 74.38 & 16.70 \\ 
    \bottomrule
  \end{tabular}}
  \label{ts_mini_hntl}
\end{table*}

\begin{table*}[!t]
\renewcommand\arraystretch{1.2}
  \centering
      \caption{The accuracy ($\%$) of target-specified SOPHON on Mini-DomainNet. The vertical/horizontal axis denotes the authorized/unauthorized domain. In each task, the left of '\(\Rightarrow\)' shows the test accuracy of SL-CLIP on the unauthorized domain, while the right side presents the accuracy of SOPHON.}
  \resizebox{1\textwidth}{!}{
  \begin{tabular}{c|cccc|cccc}
    \toprule
    Authorized/Unauthorized & Clipart & Painting & Real & Sketch & $W_{u-a} \uparrow$ & $Drop_u \uparrow$ & $Drop_a \downarrow$ \\
    \midrule   
    Clipart   & 85.1 $\Rightarrow$ 78.9 & 79.8 $\Rightarrow$ 25.2 & 89.8 $\Rightarrow$ 40.0 & 78.7 $\Rightarrow$ 25.7 & 36.55 & 52.47 & 6.17 \\ 
    Painting  & 83.8 $\Rightarrow$ 33.7 & 81.4 $\Rightarrow$ 70.2 & 89.1 $\Rightarrow$ 65.6 & 78.4 $\Rightarrow$ 22.9 & 22.38 & 43.03 & 11.17 \\ 
    Real      & 84.6 $\Rightarrow$ 36.7 & 80.5 $\Rightarrow$ 42.7 & 90.6 $\Rightarrow$ 81.4 & 80.0 $\Rightarrow$ 18.7 & 32.40 & 49.00 & 9.20 \\ 
    Sketch    & 84.3 $\Rightarrow$ 35.4 & 79.1 $\Rightarrow$ 30.6 & 90.3 $\Rightarrow$ 39.4 & 80.7 $\Rightarrow$ 72.2 & 29.55 & 49.43 & 8.50 \\ 
    \hline
    Mean      & \multicolumn{4}{c|}{/} & 30.22 & 48.48 & 8.84 \\  
    \bottomrule
  \end{tabular}}
  \label{ts_mini_sophon}
\end{table*}

\begin{table*}[!t]
\renewcommand\arraystretch{1.2}
  \centering
      \caption{The accuracy ($\%$) of target-specified IP-CLIP on Mini-DomainNet. The vertical/horizontal axis denotes the authorized/unauthorized domain. In each task, the left of '\(\Rightarrow\)' shows the test accuracy of SL-CLIP on the unauthorized domain, while the right side presents the accuracy of IP-CLIP.}
  \resizebox{1\textwidth}{!}{
  \begin{tabular}{c|cccc|cccc}
    \toprule
    Authorized/Unauthorized & Clipart & Painting & Real & Sketch & $W_{u-a} \uparrow$ & $Drop_u \uparrow$ & $Drop_a \downarrow$ \\
    \midrule   
    Clipart   & 85.1 $\Rightarrow$  84.8 & 79.8 $\Rightarrow$ 18.1 & 89.8 $\Rightarrow$  24.8 & 78.7 $\Rightarrow$  22.4 & 51.47 & 61.00 & 0.30 \\ 
    Painting  & 83.8 $\Rightarrow$ \ 9.2 & 81.4 $\Rightarrow$ 80.9 & 89.1 $\Rightarrow$  26.5 & 78.4 $\Rightarrow$  14.4 & 53.85 & 67.07 & 0.50 \\ 
    Real      & 84.6 $\Rightarrow$  21.9 & 80.5 $\Rightarrow$ 21.4 & 90.6 $\Rightarrow$  90.4 & 80.0 $\Rightarrow$ \ 6.0 & 58.82 & 65.27 & 0.20 \\
    Sketch    & 84.3 $\Rightarrow$  23.2 & 79.1 $\Rightarrow$ 15.6 & 90.3 $\Rightarrow$ \ 9.2 & 80.7 $\Rightarrow$  80.2 & 54.59 & 68.57 & 0.50 \\
    \hline
    Mean      & \multicolumn{4}{c|}{/} & 54.68& 65.48 & 0.33  \\  
    \bottomrule
  \end{tabular}}
  \label{ts_mini_ipclip}
\end{table*}

\begin{table*}[!t]
\renewcommand\arraystretch{1.2}
  \centering
      \caption{The accuracy ($\%$) of target-specified AoD-IP on Mini-DomainNet. The vertical/horizontal axis denotes the authorized/unauthorized domain. In each task, the left of '\(\Rightarrow\)' shows the test accuracy of SL-CLIP on the unauthorized domain, while the right side presents the accuracy of AoD-IP.}
  \resizebox{1\textwidth}{!}{
  \begin{tabular}{c|cccc|cccc}
    \toprule
    Authorized/Unauthorized & Clipart & Painting & Real & Sketch & $W_{u-a} \uparrow$ & $Drop_u \uparrow$ & $Drop_a \downarrow$ \\
    \midrule   
        Clipart   & 85.1 $\Rightarrow$ 85.1 & 79.8 $\Rightarrow$ 12.3 & 89.8 $\Rightarrow$ 16.3 & 78.7 $\Rightarrow$ 17.0 & 57.47 & 67.58 & 0.02 \\
        Painting  & 83.8 $\Rightarrow$ 10.4 & 81.4 $\Rightarrow$ 81.0 & 89.1 $\Rightarrow$ 17.0 & 78.4 $\Rightarrow$ 9.1 & 57.60 & 71.59 & 0.44 \\
        Real      & 84.6 $\Rightarrow$ 14.6 & 80.5 $\Rightarrow$ 14.1 & 90.6 $\Rightarrow$ 90.4 & 80.0 $\Rightarrow$ 10.7 & 61.76 & 68.58 & 0.23 \\
        Sketch    & 84.3 $\Rightarrow$ 15.9 & 79.1 $\Rightarrow$ 11.2 & 90.3 $\Rightarrow$ 12.3 & 80.7 $\Rightarrow$ 80.3 & 57.12 & 71.46 & 0.36 \\ 
    \hline
    Mean      & \multicolumn{4}{c|}{/} & 58.49 & 69.80 & 0.26 \\  
    \bottomrule
  \end{tabular}}
  \label{ts_mini_AoD-IP}
\end{table*}

\begin{table*}[!t]
\renewcommand\arraystretch{1}
\setlength{\tabcolsep}{4pt}
  \centering
      \caption{The detailed results of authorization application AoD-IP on Office-31.}
  \resizebox{1\textwidth}{!}{
  \begin{tabular}{cc|cc|cc|cc|c}
    \toprule
	$X_a$ & $A^{ip}_a \rightarrow A^{ip}_u$ & $X_{e^1}$ & $A^{ip}_{e^1} \rightarrow A^{ip}_u$ & $X_{e^2}$ & $A^{ip}_{e^2} \rightarrow A^{ip}_u$ & $R_a$ / $R_e$ / $R_u$ \\
    \midrule
    Amazon$\dagger$ & 68.1 $\rightarrow$ 4.3 / 5.1 / 4.2 & Dslr$\dagger$ & 97.7 $\rightarrow$ 20.6 / 1.2 / 0.3 & Webcam$\dagger$ & 94.7 $\rightarrow$ 28.6 / 3.2 / 9.8 & 91.5 / 94.8 / 89.4 \\
    Dslr$\dagger$ & 97.8 $\rightarrow$ 22.5 / 1.3 / 0.2 & Amazon$\dagger$ & 67.8 $\rightarrow$ 3.1 / 2.5 / 1.6 & Webcam$\dagger$ & 91.9 $\rightarrow$ 25.4 / 0.8 / 2.4 & 100.0 / 98.5 / 92.3 \\
    Webcam$\dagger$ & 92.0 $\rightarrow$ 25.8 / 1.2 / 2.4 & Amazon$\dagger$ & 68.5 $\rightarrow$ 0.8 / 1.3 / 0.6 & Dslr$\dagger$ & 97.2 $\rightarrow$ 22.1 / 1.0 / 1.3 & 96.0 / 96.4 / 93.0 \\
    \bottomrule
  \end{tabular}}
  \label{aa_office_detail}
\end{table*}

\begin{table*}[!t]
\renewcommand\arraystretch{1}
\setlength{\tabcolsep}{4pt}
  \centering
      \caption{The detailed results of authorization application AoD-IP on Mini-DomainNet.}
  \resizebox{1\textwidth}{!}{
  \begin{tabular}{cc|cc|cc|cc|c}
    \toprule
	$X_a$ & $A^{ip}_a \rightarrow A^{ip}_u$ & $X_{e^1}$ & $A^{ip}_{e^1} \rightarrow A^{ip}_u$ & $X_{e^2}$ & $A^{ip}_{e^2} \rightarrow A^{ip}_u$ & $X_{e^3}$ & $A^{ip}_{e^3} \rightarrow A^{ip}_u$ & $R_a$ / $R_e$ / $R_u$ \\
    \midrule   
    Cl$\dagger$ & 77.5 $\rightarrow$ 4.4 / 4.4 / 2.9 /  16.8 & Pa$\dagger$ & 72.4 $\rightarrow$ 4.0 / 2.4 / 3.8 /  12.4 & Re$\dagger$ & 83.4 $\rightarrow$ 1.9 / 1.7 / 1.6 /  11.6 & Sk$\dagger$ & 76.1 $\rightarrow$ 0.8 / 1.0 / 1.6 / 4.0 &  100.0 / \ 99.9 / 92.5 \\
    Pa$\dagger$ & 70.0 $\rightarrow$ 6.7 / 7.9 / 4.5 /  13.0 & Cl$\dagger$ & 77.9 $\rightarrow$ 2.9 / 1.6 / 2.9 /  20.0 & Re$\dagger$ & 83.8 $\rightarrow$ 5.7 / 4.6 / 4.6 /  13.5 & Sk$\dagger$ & 77.9 $\rightarrow$ 2.1 / 1.4 / 1.7 / 9.0 &  100.0 /  100.0 / 91.9 \\
    Re$\dagger$ & 83.2 $\rightarrow$ 5.4 / 2.4 / 1.9 /  13.0 & Cl$\dagger$ & 77.3 $\rightarrow$ 1.6 / 1.0 / 1.4 / \ 4.4 & Pa$\dagger$ & 72.8 $\rightarrow$ 5.1 / 2.7 / 3.8 /  13.3 & Sk$\dagger$ & 76.0 $\rightarrow$ 1.3 / 0.5 / 1.0 / 2.2 &  100.0 / \ 99.5 / 93.8 \\
    Sk$\dagger$ & 75.1 $\rightarrow$ 3.5 / 1.3 / 3.8 / \ 5.9 & Cl$\dagger$ & 78.9 $\rightarrow$ 2.4 / 1.4 / 3.8 / \ 7.5 & Pa$\dagger$ & 71.9 $\rightarrow$ 2.5 / 1.1 / 3.7 / \ 8.4 & Re$\dagger$ & 84.2 $\rightarrow$ 1.6 / 0.5 / 0.8 / 3.8 & \ 99.9 / \ 99.8 / 95.3 \\
    \bottomrule
  \end{tabular}}
  \label{aa_mini_detail}
\end{table*}

\begin{table*}[!t]
\renewcommand\arraystretch{1.2}
  \centering
      \caption{The results of authorization application NTL on Office-31.}
  \resizebox{0.7\textwidth}{!}{
  \begin{tabular}{c|ccc|ccc}
    \toprule
    Authorized/Test & Amazon & Dslr & Webcam & $D_{u-a} \uparrow$ & $A_u \downarrow$ & $A_a \uparrow$ \\
    \midrule 
    Amazon$\dagger$ & 57.5 & 13.8 & 41.0 & 15.67 & 37.43  & 62.50 \\
    Dslr$\dagger$   & 78.7 & 18.5 & 54.3 & 39.25 & 50.50  & 92.80 \\
    Webcam$\dagger$ & 31.8 & 17.3 & 14.8 & 54.59 & 21.30  & 85.30 \\
    \midrule 
    Mean   & \multicolumn{3}{c|}{/} & 36.50 & 36.41 & 80.20 \\
    \bottomrule
  \end{tabular}}
  \label{aa_office31_ntl}
\end{table*}

\begin{table*}[!t]
\renewcommand\arraystretch{1.2}
  \centering
      \caption{The results of authorization application CUTI-Domain on Office-31.}
  \resizebox{0.7\textwidth}{!}{
  \begin{tabular}{c|ccc|ccc}
    \toprule
    Authorized/Test & Amazon & Dslr & Webcam & $D_{u-a} \uparrow$ & $A_u \downarrow$ & $A_a \uparrow$ \\
    \midrule 
    Amazon$\dagger$ & 36.0  & \ 7.5  & 19.0 & 29.26  & 20.83  & 65.50 \\
    Dslr$\dagger$   & 74.0  & \ 6.3  & 29.3 & 54.47  & 36.53  & 94.30 \\
    Webcam$\dagger$ & 57.0  &  12.5  & 22.3 & 40.56  & 30.60  & 80.80 \\
    \midrule 
    Mean   & \multicolumn{3}{c|}{/} & 41.43  & 29.32  & 80.20 \\
    \bottomrule
  \end{tabular}}
  \label{aa_office31_cuti}
\end{table*}

\begin{table*}[!t]
\renewcommand\arraystretch{1.2}
  \centering
      \caption{The results of authorization application CUPI-Domain on Office-31.}
  \resizebox{0.7\textwidth}{!}{
  \begin{tabular}{c|ccc|ccc}
    \toprule
    Authorized/Test & Amazon & Dslr & Webcam & $D_{u-a} \uparrow$ & $A_u \downarrow$ & $A_a \uparrow$ \\
    \midrule 
    Amazon$\dagger$ & 25.1 & 7.2  & 18.5 & 31.51 & 16.92 & 65.23 \\
    Dslr$\dagger$   & 70.1 & 6.2  & 23.6 & 58.62 & 33.30 & 95.00 \\
    Webcam$\dagger$ & 45.9 & 12.3 & 18.8 & 49.70 & 25.68 & 84.50 \\
    \midrule 
    Mean   & \multicolumn{3}{c|}{/} & 46.61 & 25.30 & 81.58 \\
    \bottomrule
  \end{tabular}}
  \label{aa_office31_cupi}
\end{table*}

\begin{table*}[!t]
\renewcommand\arraystretch{1.2}
  \centering
      \caption{The results of authorization application HNTL on Office-31.}
  \resizebox{0.7\textwidth}{!}{
  \begin{tabular}{c|ccc|ccc}
    \toprule
    Authorized/Test & Amazon & Dslr & Webcam & $D_{u-a} \uparrow$ & $A_u \downarrow$ & $A_a \uparrow$ \\
    \midrule 
    Amazon$\dagger$ & 6.30 & 7.80 & 3.10 & 5.55 & 5.73 & 26.60 \\
    Dslr$\dagger$   & 46.90 & 6.30 & 10.90 & 73.19 & 21.37 & 96.90 \\
    Webcam$\dagger$ & 34.40 & 15.60 & 6.30 & 78.36 & 18.77 & 98.40 \\
    \midrule 
    Mean   & \multicolumn{3}{c|}{/} & 52.37 & 15.29 & 73.97 \\
    \bottomrule
  \end{tabular}}
  \label{aa_office31_hntl}
\end{table*}

\begin{table*}[!t]
\renewcommand\arraystretch{1.2}
  \centering
      \caption{The results of authorization application SOPHON on Office-31.}
  \resizebox{0.7\textwidth}{!}{
  \begin{tabular}{c|ccc|ccc}
    \toprule
    Authorized/Test & Amazon & Dslr & Webcam & $D_{u-a} \uparrow$ & $A_u \downarrow$ & $A_a \uparrow$ \\
    \midrule 
    Amazon$\dagger$ & 11.20 & 7.50 & 10.00 & 39.11 & 9.57 & 67.50 \\
    Dslr$\dagger$   & 7.50 & 10.00 & 12.50 & 85.31 & 10.00 & 97.50 \\
    Webcam$\dagger$ & 12.50 & 11.20 & 13.50 & 74.09 & 12.40 & 92.50 \\
    \midrule 
    Mean   & \multicolumn{3}{c|}{/} & 66.17 & 10.66 & 85.83 \\
    \bottomrule
  \end{tabular}}
  \label{aa_office31_sophon}
\end{table*}

\begin{table*}[!t]
\renewcommand\arraystretch{1.2}
  \centering
      \caption{The results of authorization application IP-CLIP on Office-31.}
  \resizebox{0.7\textwidth}{!}{
  \begin{tabular}{c|ccc|ccc}
    \toprule
    Authorized/Test & Amazon & Dslr & Webcam & $D_{u-a} \uparrow$ & $A_u \downarrow$ & $A_a \uparrow$ \\
    \midrule 
    Amazon$\dagger$ & \ 4.5 & 3.3 & \ 2.8 & 37.46 & \ 3.53 & 63.00 \\
    Dslr$\dagger$   &  27.3 & 1.5 & \ 0.5 & 82.42 & \ 9.77 & 95.80 \\
    Webcam$\dagger$ &  31.0 & 4.3 &  11.3 & 56.45 &  15.53 & 83.30 \\
    \midrule 
    Mean   & \multicolumn{3}{c|}{/} & 58.78 & \ 9.61 & 80.70 \\
    \bottomrule
  \end{tabular}}
  \label{aa_office31_ipclip}
\end{table*}

\begin{table*}[!t]
\renewcommand\arraystretch{1.2}
  \centering
      \caption{The results of authorization application AoD-IP on Office-31.}
  \resizebox{0.7\textwidth}{!}{
  \begin{tabular}{c|ccc|ccc}
    \toprule
    Authorized/Test & Amazon & Dslr & Webcam & $D_{u-a} \uparrow$ & $A_u \downarrow$ & $A_a \uparrow$ \\
    \midrule 
    Amazon$\dagger$ & 2.7 & 3.0 & 2.1 & 44.64 & 2.61 & 68.13 \\
    Dslr$\dagger$ & 21.7 & 1.2 & 0.6 & 87.55 & 7.83 & 97.57 \\
    Webcam$\dagger$ & 26.6 & 1.7 & 4.9 & 75.96 & 11.07 & 92.87 \\
    \midrule 
    Mean   & \multicolumn{3}{c|}{/} & 69.39 & \ 7.17 & 86.19 \\
    \bottomrule
  \end{tabular}}
  \label{aa_office31_AoD-IP}
\end{table*}

\begin{table*}[!t]
\renewcommand\arraystretch{1.2}
  \centering
        \caption{The results of authorization application NTL on Office-Home-65.}
  \label{aa_home_ntl}
  \resizebox{0.8\textwidth}{!}{
  \begin{tabular}{c|cccc|ccc}
    \toprule
    Authorized/Test & Art & Clipart & Product & Real-World & $D_{u-a} \uparrow$ & $A_u \downarrow$ & $A_a \uparrow$ \\
    \midrule 
    Art$\dagger$       & 21.0 & 14.3 & 17.0 & 29.5 &  49.47 & 20.45 & 81.30 \\
    Clipart$\dagger$   & 21.0 & 13.0 & 45.0 & 31.8 & \ 9.74 & 27.70 & 48.00 \\
    Product$\dagger$   & 21.3 & 27.8 & 36.8 & 24.0 &  44.44 & 27.48 & 81.80 \\
    Real-World$\dagger$ & 10.3 & 22.5 & 27.5 & 16.5 &  51.50 & 19.20 & 82.00 \\
    \midrule 
    Mean      & \multicolumn{4}{c|}{/} & 38.79 & 23.71 & 73.28 \\
    \bottomrule
  \end{tabular}}
\end{table*}

\begin{table*}[!t]
\renewcommand\arraystretch{1.2}
  \centering
        \caption{The results of authorization application CUTI-Domain on Office-Home-65.}
  \resizebox{0.8\textwidth}{!}{
  \begin{tabular}{c|cccc|ccc}
    \toprule
    Authorized/Test & Art & Clipart & Product & Real-World & $D_{u-a} \uparrow$ & $A_u \downarrow$ & $A_a \uparrow$ \\
    \midrule 
    Art$\dagger$       & \ 4.5  & \ 5.0 & 21.0 &  11.0 & 54.95 &  10.38 & 79.50 \\
    Clipart$\dagger$   &  11.0  &  16.0 & 36.5 &  20.0 & 16.86 &  20.88 & 52.80 \\
    Product$\dagger$   &  18.0  &  33.5 & 61.0 &  29.0 & 39.53 &  35.38 & 83.00 \\
    Real-World$\dagger$ & \ 7.5  & \ 3.5 & 10.8 & \ 9.5 & 62.87 & \ 7.83 & 83.30 \\
    \midrule 
    Mean      & \multicolumn{4}{c|}{/} & 43.55 & 18.61 & 74.65 \\
    \bottomrule
  \end{tabular}}
  \label{aa_home_cuti}
\end{table*}

\begin{table*}[!t]
\renewcommand\arraystretch{1.2}
  \centering
        \caption{The results of authorization application CUPI-Domain on Office-Home-65.}
  \resizebox{0.8\textwidth}{!}{
  \begin{tabular}{c|cccc|ccc}
    \toprule
    Authorized/Test & Art & Clipart & Product & Real-World & $D_{u-a} \uparrow$ & $A_u \downarrow$ & $A_a \uparrow$ \\
    \midrule 
    Art$\dagger$       & 4.1 & 5.2 & 15.1 & 9.2 & 56.53 & 8.39 & 79.50 \\
    Clipart$\dagger$   & 8.3 & 13.4 & 25.5 & 18.5 & 23.03 & 16.43 & 56.90 \\
    Product$\dagger$   & 15.1 & 21.7 & 56.8 & 27.7 & 44.41 & 30.31 & 83.50 \\
    Real-World$\dagger$ & 5.3 & 3.5 & 8.6 & 8.8 & 66.03 & 6.55 & 84.60 \\
    \midrule 
    Mean      & \multicolumn{4}{c|}{/} & 47.50 & 15.42 & 76.13 \\
    \bottomrule
  \end{tabular}}
  \label{aa_home_cupi}
\end{table*}

\begin{table*}[!t]
\renewcommand\arraystretch{1.2}
  \centering
        \caption{The results of authorization application HNTL on Office-Home-65.}
  \label{aa_home_hntl}
  \resizebox{0.8\textwidth}{!}{
  \begin{tabular}{c|cccc|ccc}
    \toprule
    Authorized/Test & Art & Clipart & Product & Real-World & $D_{u-a} \uparrow$ & $A_u \downarrow$ & $A_a \uparrow$ \\
    \midrule 
    Art$\dagger$       & 2.1 & 4.1 & 2.1 & 1.6 & 45.23 & 2.48 & 68.50 \\
    Clipart$\dagger$   & 1.6 & 3.1 & 2.1 & 1.3 & 17.62 & 2.03 & 43.00 \\
    Product$\dagger$   & 2.1 & 4.4 & 3.4 & 3.4 & 28.10 & 3.33 & 54.70 \\
    Real-World$\dagger$ & 2.6 & 3.6 & 2.1 & 3.1 & 71.00 & 2.85 & 85.70 \\
    \midrule 
    Mean      & \multicolumn{4}{c|}{/} & 40.49 & 2.67 & 62.98 \\
    \bottomrule
  \end{tabular}}
\end{table*}

\begin{table*}[!t]
\renewcommand\arraystretch{1.2}
  \centering
        \caption{The results of authorization application SOPHON on Office-Home-65.}
  \label{aa_home_sophon}
  \resizebox{0.8\textwidth}{!}{
  \begin{tabular}{c|cccc|ccc}
    \toprule
    Authorized/Test & Art & Clipart & Product & Real-World & $D_{u-a} \uparrow$ & $A_u \downarrow$ & $A_a \uparrow$ \\
    \midrule 
    Art$\dagger$       & 3.8 & 4.8 & 7.2 & 9.2 & 50.14 & 6.25 & 74.00 \\
    Clipart$\dagger$   & 5.5 & 16.2 & 13.0 & 15.2 & 26.61 & 12.48 & 58.20 \\
    Product$\dagger$   & 2.0 & 3.8 & 10.2 & 4.5 & 64.64 & 5.13 & 83.00 \\
    Real-World$\dagger$ & 4.8 & 4.0 & 13.2 & 10.0 & 59.44 & 8.00 & 81.20 \\
    \midrule 
    Mean      & \multicolumn{4}{c|}{/} & 50.21 & 7.96 & 74.10 \\
    \bottomrule
  \end{tabular}}
\end{table*}

\begin{table*}[!t]
\renewcommand\arraystretch{1.2}
  \centering
        \caption{The results of authorization application IP-CLIP on Office-Home-65.}
  \resizebox{0.8\textwidth}{!}{
  \begin{tabular}{c|cccc|ccc}
    \toprule
    Authorized/Test & Art & Clipart & Product & Real-World & $D_{u-a} \uparrow$ & $A_u \downarrow$ & $A_a \uparrow$ \\
    \midrule 
    Art       & 1.5 & 3.3 & \ 7.8 & 3.0 & 60.12 & \ 3.88 & 79.50 \\
    Clipart   & 4.3 & 5.3 &  22.5 & 9.8 & 26.52 &  10.48 & 57.00 \\
    Product   & 5.8 & 9.3 &  12.0 & 6.5 & 57.74 & \ 8.40 & 80.30 \\
    Real-World & 2.3 & 4.0 & \ 8.5 & 6.0 & 71.17 & \ 5.20 & 87.00 \\
    \midrule 
    Mean      & \multicolumn{4}{c|}{/} & 53.89 & \ 6.99 & 75.95 \\
    \bottomrule
  \end{tabular}}
  \label{aa_home_ipclip}
\end{table*}

\begin{table*}[!t]
\renewcommand\arraystretch{1.2}
  \centering
        \caption{The results of authorization application AoD-IP on Office-Home-65.}
  \resizebox{0.8\textwidth}{!}{
  \begin{tabular}{c|cccc|ccc}
    \toprule
    Authorized/Test & Art & Clipart & Product & Real-World & $D_{u-a} \uparrow$ & $A_u \downarrow$ & $A_a \uparrow$ \\
    \midrule 
    Art$\dagger$       & 1.6 & 1.8 & 8.6 & 2.9 & 64.29 & 3.11 & 81.75 \\
    Clipart$\dagger$   & 2.8 & 4.9 & 15.7 & 5.9 & 28.95 & 5.28 & 56.51 \\
    Product$\dagger$   & 1.7 & 2.6 & 8.3 & 2.4 & 67.55 & 3.73 & 84.08 \\
    Real-World$\dagger$ & 1.7 & 2.6 & 6.0 & 2.3 & 73.77 & 3.14 & 87.48 \\
    \midrule 
    Mean      & \multicolumn{4}{c|}{/} & 58.64 & 3.82 & 77.45 \\
    \bottomrule
  \end{tabular}}
  \label{aa_home_AoD-IP}
\end{table*}

\begin{table*}[!t]
\renewcommand\arraystretch{1.2}
  \centering
        \caption{The results of authorization application NTL on Mini-DomainNet.}
  \resizebox{0.8\textwidth}{!}{
  \begin{tabular}{c|cccc|ccc}
    \toprule
    Authorized/Test & Clipart & Painting & Real & Sketch & $D_{u-a} \uparrow$ & $A_u \downarrow$ & $A_a \uparrow$ \\
    \midrule 
    Clipart$\dagger$   & 13.5 & \ 8.7 & 12.1 & 35.9 & 38.45 & 17.54 & 71.40 \\
    Painting$\dagger$  & 26.2 &  15.1 & 14.3 & 41.1 & 32.78 & 24.18 & 70.60 \\
    Real$\dagger$      & 41.1 &  21.8 & 24.6 & 64.1 & 35.66 & 37.90 & 81.60 \\
    Sketch$\dagger$    & 16.5 & \ 7.8 & 11.3 & 30.6 & 38.66 & 16.55 & 71.00 \\
    \midrule 
    Mean      & \multicolumn{4}{c|}{/} & 36.39 & 24.04 & 73.65 \\       
    \bottomrule
  \end{tabular}}
  \label{aa_mini_ntl}
\end{table*}

\begin{table*}[!t]
\renewcommand\arraystretch{1.2}
  \centering
        \caption{The results of authorization application CUTI-Domain on Mini-DomainNet.}
  \resizebox{0.8\textwidth}{!}{
  \begin{tabular}{c|cccc|ccc}
    \toprule
    Authorized/Test & Clipart & Painting & Real & Sketch & $D_{u-a} \uparrow$ & $A_u \downarrow$ & $A_a \uparrow$ \\
    \midrule 
    Clipart$\dagger$   &  57.9 &  24.3 &  31.1 & 63.0 & 22.77 &  44.08 & 74.60 \\
    Painting$\dagger$  &  46.5 &  13.8 &  21.6 & 57.0 & 24.48 &  34.73 & 69.80 \\
    Real$\dagger$      &  42.4 &  17.9 &  21.9 & 57.1 & 33.56 &  34.83 & 77.90 \\
    Sketch$\dagger$    & \ 6.0 & \ 4.8 & \ 6.4 & 20.6 & 48.18 & \ 9.45 & 74.30 \\
    \midrule 
    Mean      & \multicolumn{4}{c|}{/} & 32.25 & 30.77 & 74.15 \\
    \bottomrule
  \end{tabular}}
  \label{aa_mini_cuti}
\end{table*}

\begin{table*}[!t]
\renewcommand\arraystretch{1.2}
  \centering
        \caption{The results of authorization application CUPI-Domain on Mini-DomainNet.}
  \resizebox{0.8\textwidth}{!}{
  \begin{tabular}{c|cccc|ccc}
    \toprule
    Authorized/Test & Clipart & Painting & Real & Sketch & $D_{u-a} \uparrow$ & $A_u \downarrow$ & $A_a \uparrow$ \\
    \midrule 
    Clipart$\dagger$   & 30.5 & 18.9 & 26.7 & 40.7 & 34.35 & 29.21 & 75.00 \\
    Painting$\dagger$  & 30.7 & 12.1 & 16.7 & 18.2 & 35.64 & 19.43 & 70.20 \\
    Real$\dagger$      & 29.4 & 15.1 & 18.3 & 37.5 & 44.35 & 25.08 & 80.30 \\
    Sketch$\dagger$    & 6.5 & 4.2 & 5.8 & 15.6 & 50.24 & 8.02 & 75.00 \\
    \midrule 
    Mean      & \multicolumn{4}{c|}{/} & 41.14 & 20.43 & 75.13 \\
    \bottomrule
  \end{tabular}}
  \label{aa_mini_cupi}
\end{table*}

\begin{table*}[!t]
\renewcommand\arraystretch{1.2}
  \centering
        \caption{The results of authorization application HNTL on Mini-DomainNet.}
  \resizebox{0.8\textwidth}{!}{
  \begin{tabular}{c|cccc|ccc}
    \toprule
    Authorized/Test & Clipart & Painting & Real & Sketch & $D_{u-a} \uparrow$ & $A_u \downarrow$ & $A_a \uparrow$ \\
    \midrule 
    Clipart$\dagger$   & 1.8 & 0.3 & 1.6 & 0.8 & 2.60 & 1.11 & 16.70 \\
    Painting$\dagger$  & 0.3 & 0.3 & 0.5 & 0.3 & 0.47 & 0.35 & 7.00 \\
    Real$\dagger$      & 1.0 & 0.5 & 1.0 & 1.3 & 2.05 & 0.95 & 14.80 \\
    Sketch$\dagger$    & 0.5 & 0.8 & 1.3 & 1.3 & 0.42 & 0.98 & 7.00 \\
    \midrule 
    Mean      & \multicolumn{4}{c|}{/} & 1.39 & 0.85 & 11.38 \\
    \bottomrule
  \end{tabular}}
  \label{aa_mini_hntl}
\end{table*}

\begin{table*}[!t]
\renewcommand\arraystretch{1.2}
  \centering
        \caption{The results of authorization application SOPHON on Mini-DomainNet.}
  \resizebox{0.8\textwidth}{!}{
  \begin{tabular}{c|cccc|ccc}
    \toprule
    Authorized/Test & Clipart & Painting & Real & Sketch & $D_{u-a} \uparrow$ & $A_u \downarrow$ & $A_a \uparrow$ \\
    \midrule 
    Clipart$\dagger$   & 24.1 & 5.1 & 8.4 & 15.9 & 50.55 & 13.38 & 78.10 \\
    Painting$\dagger$  & 16.0 & 10.6 & 11.4 & 14.6 & 41.59 & 13.15 & 71.40 \\
    Real$\dagger$      & 11.7 & 6.3 & 9.4 & 4.4 & 60.10 & 7.95 & 81.60 \\
    Sketch$\dagger$    & 10.0 & 5.9 & 7.0 & 3.7 & 45.15 & 6.65 & 70.60 \\
    \midrule 
    Mean      & \multicolumn{4}{c|}{/} & 49.35 & 10.28 & 75.43 \\
    \bottomrule
  \end{tabular}}
  \label{aa_mini_sophon}
\end{table*}

\begin{table*}[!t]
\renewcommand\arraystretch{1.2}
  \centering
        \caption{The results of authorization application IP-CLIP on Mini-DomainNet.}
  \resizebox{0.8\textwidth}{!}{
  \begin{tabular}{c|cccc|ccc}
    \toprule
    Authorized/Test & Clipart & Painting & Real & Sketch & $D_{u-a} \uparrow$ & $A_u \downarrow$ & $A_a \uparrow$ \\
    \midrule 
    Clipart$\dagger$   & \ 4.6 & 5.1 & 4.8 & 14.9 & 50.88 & \ 7.35 & 75.10 \\
    Painting$\dagger$  & \ 7.5 & 4.6 & 8.4 & 23.2 & 40.33 &  10.93 & 69.20 \\
    Real$\dagger$      &  17.3 & 8.4 & 9.8 & 38.1 & 54.06 &  18.40 & 83.30 \\
    Sketch$\dagger$    & \ 4.6 & 3.8 & 5.7 & 18.7 & 48.27 & \ 8.20 & 73.70 \\
    \midrule 
    Mean      & \multicolumn{4}{c|}{/} & 48.39 &  11.22 & 75.33 \\
    \bottomrule
  \end{tabular}}
  \label{aa_mini_ipclip}
\end{table*}

\begin{table*}[!t]
\renewcommand\arraystretch{1.2}
  \centering
        \caption{The results of authorization application AoD-IP on Mini-DomainNet.}
  \resizebox{0.8\textwidth}{!}{
  \begin{tabular}{c|cccc|ccc}
    \toprule
    Authorized/Test & Clipart & Painting & Real & Sketch & $D_{u-a} \uparrow$ & $A_u \downarrow$ & $A_a \uparrow$ \\
    \midrule 
    Clipart$\dagger$   & 2.8 & 2.1 & 2.8 & 12.2 & 56.82 & 4.96 & 77.90 \\
    Painting$\dagger$  & 4.6 & 3.5 & 4.0 & 11.8 & 47.24 & 5.96 & 71.78 \\
    Real$\dagger$      & 3.7 & 2.3 & 2.2 & 10.5 & 66.07 & 4.66 & 83.65 \\
    Sketch$\dagger$    & 1.9 & 1.1 & 2.0 & 5.3  & 56.22 & 2.57 & 76.28 \\
    \midrule 
    Mean & \multicolumn{4}{c|}{/} & 56.59 & 4.54 & 77.40 \\
    \bottomrule
  \end{tabular}}
  \label{aa_mini_AoD-IP}
\end{table*}

\begin{table*}[!t]
\renewcommand\arraystretch{1.6}
  \centering
      \caption{Ablation study of target-specified AoD-IP 'Cl $\rightarrow$ Pa'.}

  \resizebox{0.8\textwidth}{!}{
    \begin{tabular}{c|cccccc}
    \toprule
    $\lambda_1$ & 0.01 & 0.05 & 0.1 & 0.2 & 0.5 & 1 \\
    \midrule
    $X_a \rightarrow X_u$ & 84.1 $\rightarrow$ 12.8 & 83.9 $\rightarrow$ 13.0 & 84.0 $\rightarrow$ 11.9 & 82.8 $\rightarrow$ 12.5 & 83.2 $\rightarrow$ 11.8 & 93.6 $\rightarrow$ 12.1 \\
    \bottomrule
\end{tabular}}
\label{hyperparameter}
\end{table*}

\begin{table*}[!t]
\renewcommand\arraystretch{1.6}
  \centering
    \caption{A detailed description of the augmentation method applied in the target-free scenario.}
  \resizebox{0.8\textwidth}{!}{
  \begin{tabular}{cc}
    \toprule
    Augmentation & Augmentation\\

    \midrule 
    AutoContrast   & Applies automatic contrast adjustment to an image. \\
    Brightness     & Adjusts the brightness of an image. \\
    Color          & Adjusts the color saturation of an image. \\
    Contrast       & Adjusts the contrast of an image. \\
    Equalize       & Equalizes the histogram of an image. \\
    Identity       & Returns the image without any changes. \\
    Posterize      & Reduces the color depth of an image. \\
    Rotate         & Rotates an image by a random degree. \\
    Sharpness      & Adjusts the sharpness of an image. \\
    ShearX         & Shears an image along the X-axis. \\
    ShearY         & Shears an image along the Y-axis. \\
    Solarize       & Inverts all pixel values above a threshold \\
    TranslateX     & Translates an image horizontally. \\
    TranslateY     & Translates an image vertically \\
    \bottomrule
  \end{tabular}}
  \label{aug}
\end{table*}

\end{document}